\documentclass[12pt, draftclsnofoot, journal, letterpaper, onecolumn]{IEEEtran}

\makeatletter
\def\ps@headings{%
\def\@oddhead{\mbox{}\scriptsize\rightmark \hfil \thepage}%
\def\@evenhead{\scriptsize\thepage \hfil \leftmark\mbox{}}%
\def\@oddfoot{}%
\def\@evenfoot{}}
\makeatother 

\IEEEoverridecommandlockouts

\usepackage{amsfonts}
\usepackage[dvips]{graphicx}
\usepackage{caption}
\usepackage{subfigure}
\usepackage{times}
\usepackage{cite}
\usepackage{lettrine}
\usepackage{amsmath}
\usepackage{amsmath}
\usepackage{amsthm}
\allowdisplaybreaks[4]
\usepackage{array}
\usepackage{amssymb}

\usepackage{stfloats}
\usepackage{slashbox}
\usepackage{graphicx}
\usepackage{footnote}
\usepackage{booktabs}
\usepackage{array}
\usepackage{algorithmic}
\usepackage{algorithm}
\usepackage{subeqnarray}
\usepackage{cases}
\usepackage{threeparttable}
\usepackage{color}

\newcommand{\bm}[1]{\mbox{\boldmath{$#1$}}}

\UseRawInputEncoding
\begin{document}

\title{Optimal Model Placement and Online Model Splitting for Device-Edge Co-Inference}

\IEEEoverridecommandlockouts

\author{Jia~Yan,~\IEEEmembership{Student Member,~IEEE}, Suzhi~Bi,~\IEEEmembership{Senior Member,~IEEE}, and Ying-Jun~Angela~Zhang,~\IEEEmembership{Fellow,~IEEE}
\thanks{J. Yan (yj117@ie.cuhk.edu.hk) and Y. J. Zhang (yjzhang@ie.cuhk.edu.hk) are with the Department of Information Engineering, The Chinese University of Hong Kong, Hong Kong. S. Bi (bsz@szu.edu.cn) is with the College of Electronic and Information Engineering, Shenzhen University, Shenzhen, China.}
}
%\author{Jia ~Yan$^\dagger$, Suzhi~Bi$^*$, and Ying-Jun~Angela~Zhang$^\dagger$\\
%$^\dagger$Department of Information Engineering, The Chinese University of Hong Kong, Shatin, N.T., Hong Kong SAR\\
%$^*$College of Information Engineering, Shenzhen University, Shenzhen, Guangdong, China 518060\\
%E-mail:~ \{yj117, yjzhang\}@ie.cuhk.edu.hk, ~bsz@szu.edu.cn \vspace{-2ex}}

\maketitle

\vspace{-1.5cm}

\begin{abstract}
   Device-edge co-inference opens up new possibilities for resource-constrained wireless devices (WDs) to execute deep neural network (DNN)-based applications with heavy computation workloads. In particular, the WD executes the first few layers of the DNN and sends the intermediate features to the edge server that processes the remaining layers of the DNN. By adapting the model splitting decision, there exists a tradeoff between local computation cost and communication overhead. In practice, the DNN model is re-trained and updated periodically at the edge server. Once the DNN parameters are regenerated, part of the updated model must be placed at the WD to facilitate on-device inference. In this paper, we study the joint optimization of the model placement and online model splitting decisions to minimize the energy-and-time cost of device-edge co-inference in presence of wireless channel fading. The problem is challenging because the model placement and model splitting decisions are strongly coupled, while involving two different time scales. We first tackle online model splitting by formulating an optimal stopping problem, where the finite horizon of the problem is determined by the model placement decision. In addition to deriving the optimal model splitting rule based on backward induction, we further investigate a simple one-stage look-ahead rule, for which we are able to obtain analytical expressions of the model splitting decision. The analysis is useful for us to efficiently optimize the model placement decision in a larger time scale. In particular, we obtain a closed-form model placement solution for the fully-connected multilayer perceptron with equal neurons.
   %Furthermore, we propose a hybrid algorithm to balance between solution optimality and computational complexity.
   Simulation results validate the superior performance of the joint optimal model placement and splitting with various DNN structures.

\end{abstract}

%In device-edge co-inference, the wireless device (WD)  executes the first few layers of the deep neural network (DNN) and sends the intermediate features to the edge server that processes the remaining layers of the DNN. In this paper, we study the joint optimization of the model placement and online model splitting decisions to minimize the energy-and-time cost of device-edge co-inference in presence of wireless channel fading. The problem is challenging because the model placement and model splitting decisions are strongly coupled, while involving two different time scales. We first tackle online model splitting by formulating an optimal stopping problem, where the finite horizon of the problem is determined by the model placement decision. In addition to deriving the optimal model splitting rule based on backward induction, we further investigate a simple one-stage look-ahead rule, for which we are able to obtain analytical expressions of the model splitting decision. The analysis is useful to efficiently optimize the model placement decision in a larger time scale. In particular, we obtain a closed-form model placement solution for the fully-connected multilayer perceptron with equal neurons. Simulation results validate the superior performance of the joint optimal model placement and splitting with various DNN structures.

\section{Introduction}

\subsection{Motivation and Contributions}

With recent advancements in artificial intelligence (AI) \cite{eisurvey,ai3}, many deep neural network (DNN)-based applications have emerged in mobile systems \cite{ai1,ai2}, such as human face recognition and augmented reality. Due to the tremendous amount of computation workload, the DNN-based applications cannot be fully executed at the wireless devices (WDs) with low-performance computing units and limited battery life \cite{ondevice1,ondevice2,Bi_service,Jia_pricing,Zehong}. Alternatively, the WDs can choose to offload the computations to a nearby server located at the network edge, referred to as edge inference \cite{eisurvey}. Typically, the edge server can execute the whole DNN-based application on the WD's behalf after receiving the raw input data from the WD. However, due to the massive original input data (e.g., 3D images and videos), the excessive communication overhead makes it impractical to support delay-sensitive services \cite{serverbased1,serverbased2}.  Such difficulty can be overcome by performing \emph{device-edge co-inference}, where a large DNN is splitted into two parts. The first part with computation-friendly workload is executed on the WD, while the remaining part is computed on the edge server.  The WD needs to transmit the output of the first part (i.e., the intermediate feature) to the edge server for further execution.

It is essential to determine at which layer the WD splits the DNN model, i.e., stops local computing and offloads the intermediate feature.  The prior work on model splitting \cite{ei1,ei2,ei3,ei4,ei5,ei6} showed that by carefully selecting the model splitting point, one can strike a balance between the on-device computation workload and the offloading communication overhead. Take the AlexNet \cite{alexnet} for example. Fig. 1 shows that a deeper splitting point (i.e., splitting the DNN at its latter layer) in the AlexNet leads to a larger local computation workload and lower offloading data size. Besides, the model splitting decisions are affected by the time-varying wireless channel fading, e.g., deep fading may lead to a large offloading cost from the WD to the edge server. The existing model splitting methods \cite{ei2,ei3,ei4,ei5,ei6} are based on offline optimization assuming non-causal channel knowledge.
%The related work applies either manual selection \cite{ei2,ei3,ei5} or exhaustive search method \cite{ei4,ei6}, based on the assumption that the WD knows the channel gains at all the feasible splitting points of the DNN.
However, in practice, it is difficult for a WD to predict the channel state information (CSI) at a forthcoming model splitting point. Therefore, the model splitting point selection is an online decision process, where decisions must be made based on the past and present observations of wireless channel conditions without any future channel knowledge.

\begin{figure}[htb]
\begin{centering}
\includegraphics[scale=0.5]{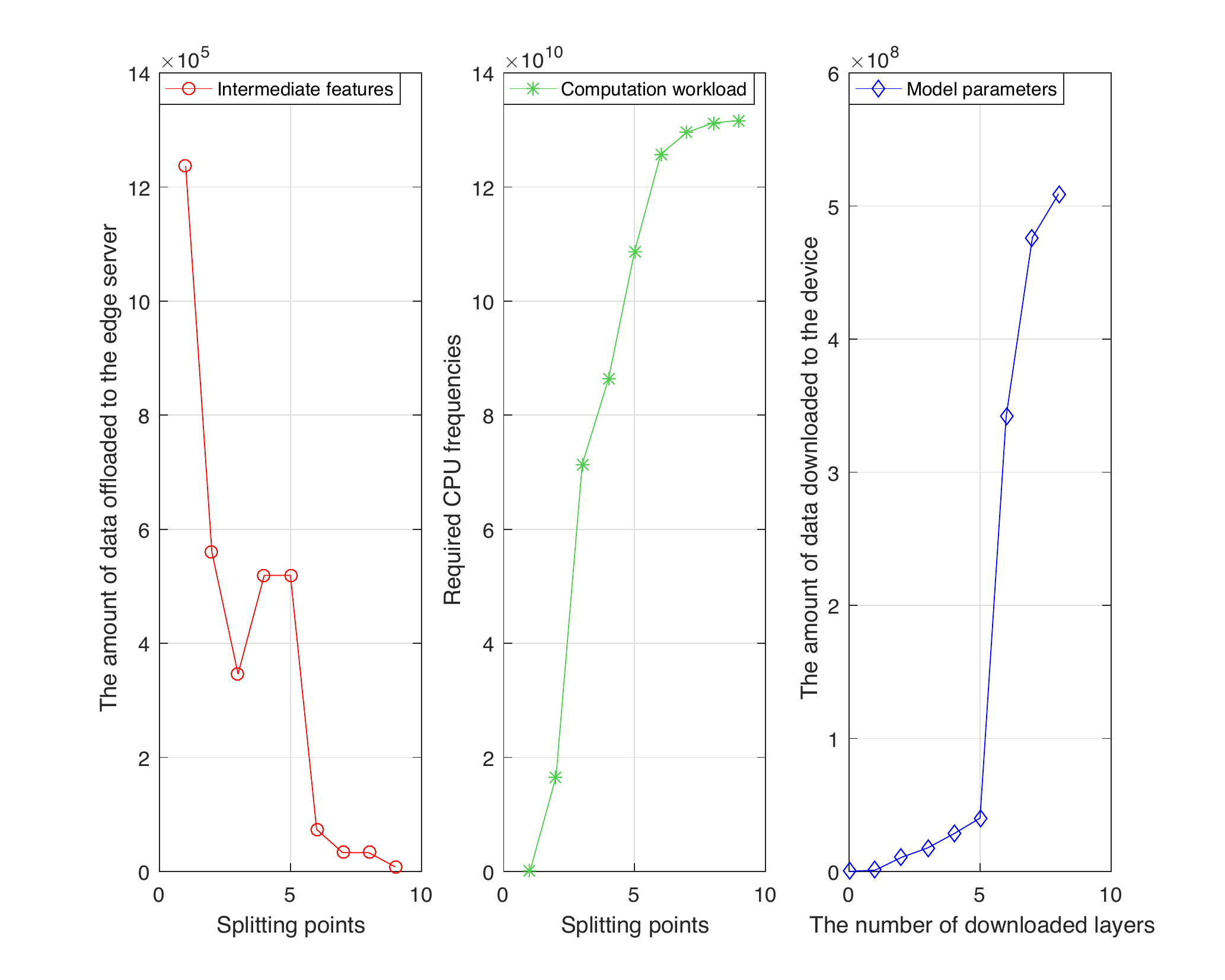}
\vspace{-0.1cm}
 \caption{The intermediate feature size, computation workload and model parameter data size in the AlexNet. }
\end{centering}
\vspace{-0.1cm}
\end{figure}

Prior work on device-edge co-inference assumes that the DNN model is completely stored at the WD.  In practice, the DNN model needs to be updated from time to time through a training process at the edge server \cite{eisurvey,ai3,alexnet}. Once the model parameters are regenerated, the WD needs to download them from the edge server to facilitate on-device inference. Such model parameter downloading process is time-consuming in a wireless system due to the large model parameter size. For example, the total model parameter size in bytes is in the order of $10^8$ in the AlexNet. We argue that for device-edge co-inference, it is not necessary to place all the DNN layers at the WD. This is because the layers after the model splitting point are never executed at the WD. Noticeably, the model placement decision must be jointly optimized with the model splitting strategy, because the WD is not able to split at a layer that is not downloaded.

%Accordingly, by taking the model downloading cost into account, it is not necessary to place all the layers of the DNN at the WD for device-edge co-inference. This is because unavailable model parameters of the layers behind the model splitting point do not affect the on-device inference performance while reducing the model downloading cost. As such, we should determine how many the early layers of the DNN to be placed at the WD, which is highly related to the model splitting point selection.

The model placement and splitting decisions are made on two different time scales. On one hand, we need to make the model splitting decision for every DNN inference process due to the fast variation of wireless channel conditions. On the other hand, the DNN model parameters are updated at a much lower frequency than the DNN inference requests, and thus the model placement decision is made on a much larger time scale \cite{eisurvey,ai3,alexnet}.

In this paper, we are interested in answering the following two key questions:
\begin{enumerate}
  \item \emph{On a large time scale, how many layers of the DNN model shall be placed at the WD, so that the expected device-edge co-inference cost is minimized?}
  \item \emph{On a fast time scale, how to choose the model splitting point to achieve the optimal tradeoff between the on-device computation and communication overhead when the future CSI is unknown?}
\end{enumerate}

%The challenges for jointly addressing the above two questions are twofold.
%First, the model placement decision provides a constraint for choosing the model splitting points. That is, the WD cannot decouple the DNN at the layers that are not locally deployed. Second, the model placement solution highly depends on the long-term performance of the online model splitting strategy. That is, in order to reduce the model downloading cost, the WD tends to store only those layers that are in front of the model splitting point with a high probability.

%We consider in this paper, a device-edge co-inference system with one WD and one edge server executing a DNN-based application. In particular, we intend to minimize the average energy and delay cost of executing the DNN-based application for $K \geq 1$ times. To the best of our knowledge, this is the first work to jointly optimize the model placement decision and online model splitting strategy of device-edge co-inference in fading channels.
%This paper  endeavors to address the above mentioned challenges for device-edge co-inference in mobile systems with wireless channel fading.

The main contributions of this paper are summarized in the following.
\begin{itemize}
  \item \emph{Online Model Splitting Strategy:} For given model placement decision, we formulate the optimal model splitting point selection problem as an optimal stopping problem \cite{stopping,stopping1,stopping2,stopping3} with finite horizon. We then solve the optimal stopping problem by backward induction to find the optimal model splitting strategy (i.e., the optimal stopping rule). Besides, we propose a suboptimal one-stage look-ahead (1-sla) stopping rule, where the analytical expressions of the model splitting strategy can be derived. Accordingly, we further analyze the optimality probability of the 1-sla stopping rule.
  \item \emph{Optimal Model Placement Algorithm:} Based on the optimal stopping rule, we derive the long-term expected cost of the WD as a function of the model placement decision. Accordingly, the optimal model placement can be obtained by enumerating the $N+1$ possible decisions, where $N$ is the total number of layers of the DNN. The brute-force search based model placement algorithm is computationally expensive, mainly because evaluating each model placement decision involves the full process of backward induction.  To reduce the complexity, we propose an efficient 1-sla stopping rule based model placement algorithm.
      %Besides, by exploring the monotone nature, we theoretically analyze the optimality probability of the 1-sla stopping rule. Based on the 1-sla stopping rule, we propose an efficient algorithm to optimize the model placement decision.
      In particular, for a fully-connected multilayer perceptron with equal neurons at all the layers, we show that the optimal model placement solution can be obtained in closed form.
  \item \emph{Performance Improvement:} Our simulation results show that the optimality probability of 1-sla model splitting strategy is as high as $0.9$ in the AlexNet for any model placement decision. Besides, we demonstrate that the joint model placement and splitting algorithm significantly reduces the overall device-edge co-inference cost under various DNN structures.

  %\item \emph{An Alternative to Balance Between Solution Quality and Computational Complexity:} We further propose a hybrid algorithm that first applies the proposed 1-sla stopping rule to find the model placement decision and then uses backward induction method to find the optimal model splitting strategy. Our simulation results show that under various DNN structures, the proposed hybrid algorithm enhances the overall device-edge co-inference performance compared with the 1-sla stopping rule based method in compensation for only a slight increase of the computational complexity.
\end{itemize}

%Simulation results show that under various DNN structures, our proposed 1-sla stopping rule based method and the hybrid algorithm achieve the near-optimal performances, while reducing the computational complexity compared to the optimal backward induction based searching method.

\subsection{Related Work}

Existing work has extensively investigated the model splitting problem for device-edge co-inference \cite{ei1,ei2,ei3,ei4,ei5,ei6}. Specifically, \cite{ei3} proposed a three-step framework including the model splitting point selection, the communication-aware on-device model compression, and the task-oriented encoding for intermediate features. The model splitting point is selected via exhaustive search therein. The authors in \cite{ei1} formulated the model splitting problem as an integer linear programming problem. In \cite{ei4}, a lightweight scheduler was designed to partition the DNN. \cite{ei5} investigated the encoding of the feature space for energy saving in edge-host partitioning of DNN. The authors in \cite{ei6} proposed a 2-step pruning framework for DNN splitting in device-edge co-inference. The key assumption in \cite{ei1,ei2,ei3,ei4,ei5,ei6} is that the WD knows the CSI at all the possible splitting points beforehand. This assumption, however, does not hold in practice since the wireless channel conditions at the forthcoming model decoupling points are random and unknown a priori.

%The theory of optimal stopping \cite{stopping} has been widely studied in the wireless communication systems, e.g., the wireless random-access networks \cite{stopping1}, the wireless power transfer based mobile edge computing \cite{stopping2}, and the cognitive radio \cite{stopping3}. Specifically, \cite{stopping3} investigated the optimal spectrum sensing decision in the cognitive radio networks and formulated it as an optimal stopping problem. To our best knowledge, our work in this paper is a first attempt to introduce the optimal stopping theory to the online model splitting point selection in edge inference.

Besides, the existing work \cite{ei1,ei2,ei3,ei4,ei5} assumes that the WD has already stored the whole DNN model to enable device-edge co-inference. This incurs significant model placement cost when the DNN model is frequently updated.
%Due to the periodical update property of the DNN, the model stored at the WD is not always valid.
In \cite{ei6}, the WD downloads only part of the model that is needed for device-side computation.  Nevertheless, the model placement decision is pre-determined and not optimized therein.  In this regard, the joint optimization of the model placement and online model partition strategy is largely overlooked in the literatures. This paper is among the first attempts to fill this gap.

\subsection{Organization}

The rest of the paper is organized as follows. Section II introduces the system model and the problem formulation. The optimal model splitting and placement strategies based on backward induction are proposed in Section III. In Section IV, we propose reduced-complexity algorithms based on the 1-sla stopping rule. We further propose a hybrid algorithm to balance the solution optimality and the computational complexity in Section IV. In Section V, simulation results are described. Finally, we conclude the paper in Section VI.

\section{System Model and Problem Formulation}

\begin{figure}[htb]
\begin{centering}
\includegraphics[scale=0.7]{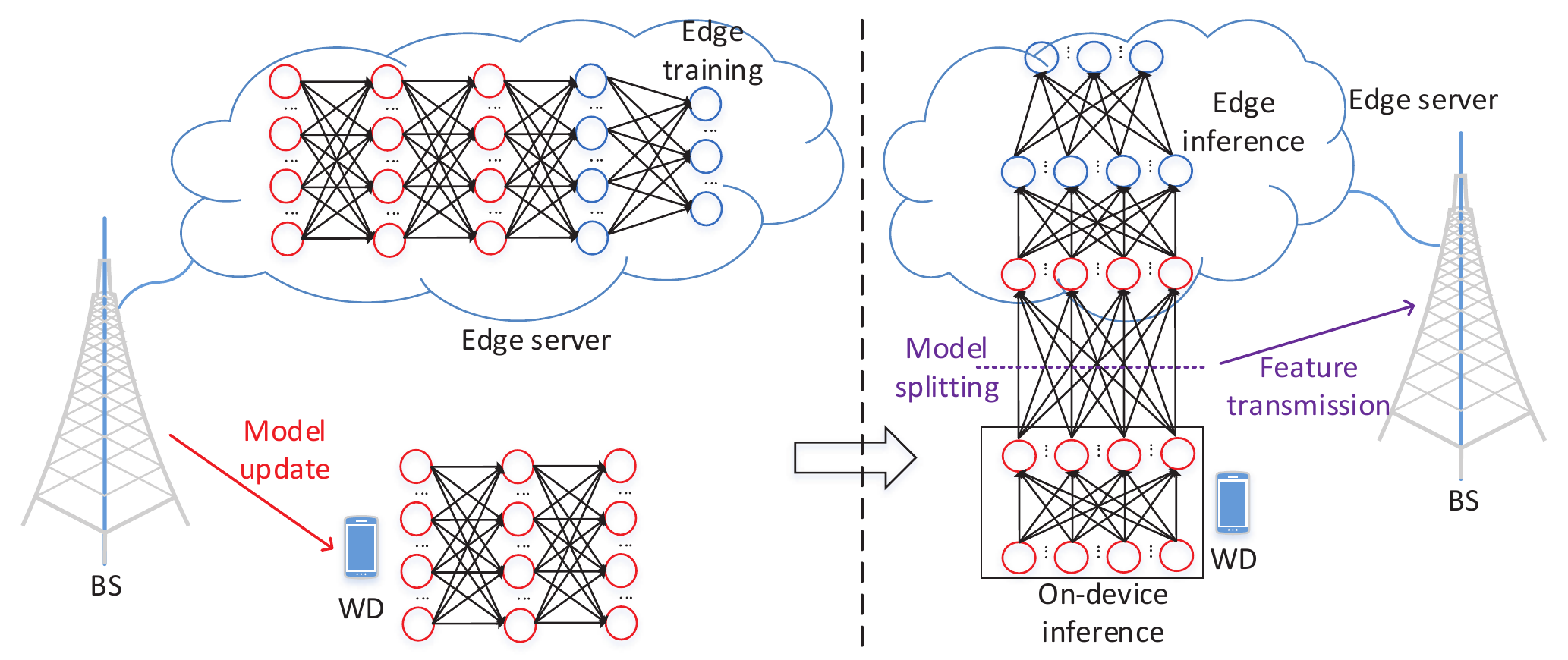}
\vspace{-0.1cm}
 \caption{An illustration of the model placement and model splitting for device-edge co-inference. }\label{fig2}
\end{centering}
\vspace{-0.1cm}
\end{figure}

\subsection{System Model}

As shown in Fig. \ref{fig2}, we consider a mobile edge inference system with one base station (BS) and one WD. The BS is the gateway of the edge server and has a stable power supply. We consider a DNN-based application with the layered network structure, where the parameters of the DNN need to be periodically updated through a training process at the edge server. The edge server is interested in the inference result, i.e., the output of the DNN. On the other hand, the input of the DNN (e.g., image for the DNN-based human face recognition) is generated by the WD.

In most existing implementation, edge inference is either executed on device (device-only inference) or fully offloaded to the edge server (edge-only inference). To strike a balance between computation and communication overhead, we consider a device-edge co-inference framework. The WD executes the first few layers of the DNN and forwards the intermediate features to the edge server that executes the remaining layers. To enable device-edge co-inference, the BS needs to place the first few layers of the DNN at the WD. Note that once the DNN model is updated, the BS shall re-send the model to the WD.
Take Fig. \ref{fig2} for example. The BS deploys the first 3 layers (marked in red) at the WD. Then, the WD can choose to split the DNN (i.e., stop local execution and offload the intermediate features) after computing 0, 1, 2 or all the 3 layers.

%Notice that the DNN needs to be periodically trained at the edge server. Once the DNN model is updated, the edge server will deploy some of its early layers to the WD.
Denote by $N$ the total number of layers of the DNN, and by $M, 0\leq M\leq N$, the number of layers placed at the WD. Suppose that the model can be used for $K$ inference tasks before it is updated. Suppose that the downloading time of the $i$-th DNN layer is $\tau_i^m$.
%%
%\begin{align}
%\tau_i^m=\frac{c_i}{R^m}.
%\end{align}
%%
Then, the average model placement cost per inference task is
\begin{align}\label{download}
\psi(M)=\frac{\sum_{i=1}^{M}\tau_i^m}{K}.
\end{align}

\begin{figure}[htb]
\begin{centering}
\includegraphics[scale=0.5]{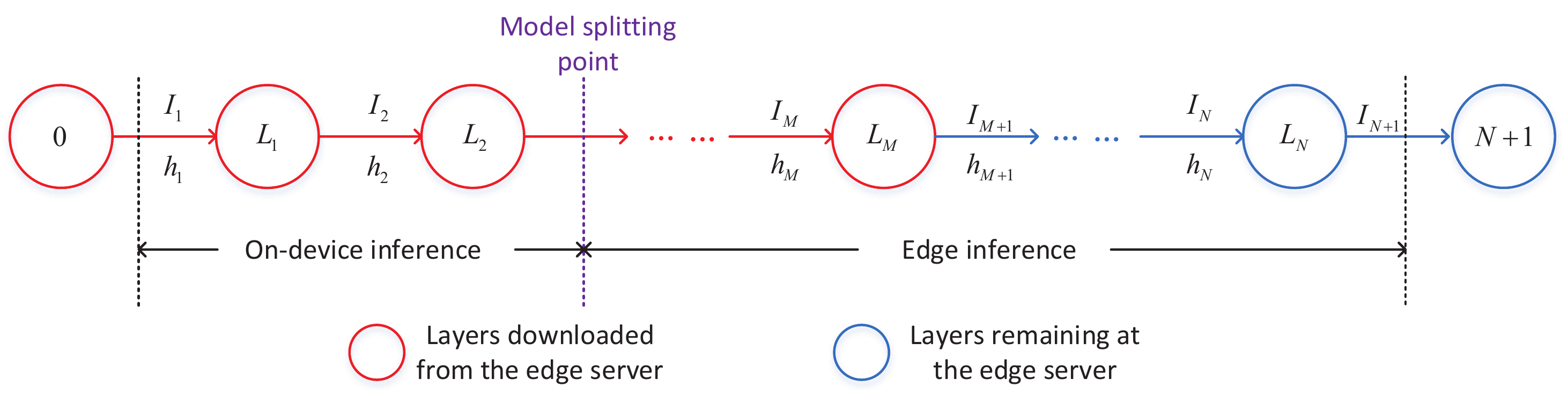}
\vspace{-0.1cm}
 \caption{Representation of the DNN as a sequential task graph. }\label{fig1}
\end{centering}
\vspace{-0.1cm}
\end{figure}

In Fig. \ref{fig1}, we model the DNN based inference by a sequential task graph. Each vertex in the task graph represents a subtask, i.e., one layer of the DNN.
We denote the computational workload of subtask $i$ in terms of the total number of CPU cycles as $L_i$.
Besides, each edge in the task graph represents that the input data of subtask $i$ is the output of the preceding subtask $i-1$. We denote the input data size in bits of subtask $i$ as $I_i$. To reflect the fact that the input data is originated from the WD and the inference output is required by the edge server, we introduce two virtual subtasks $0$ and $N+1$ as the entry and exit subtasks, respectively. In particular, subtasks $0$ and $N+1$ must be executed at the WD and the edge server, respectively. Specifically, $L_0=L_{N+1}=0$.

We define the model splitting point $n_k$ of inference task $k, k=1,...,K,$ if subtasks $0$ to $n_k-1$ are executed on the WD and subtasks $n_k$ to $N+1$ are computed at the edge server. The model splitting point $n_k$ is determined by balancing the local computing cost, uplink transmission cost of the input data $I_{n_k}$ of subtask $n_k$, and the edge computing cost. In the following, we focus on a tagged inference task and drop the subscript $k$ for notational brevity. Note that the model splitting point is constrained by the number of layers placed at the WD. That is,
\begin{align}
1\leq n\leq M+1.
\end{align}
In particular, $n=1$ implies edge-only inference, while $n=N+1$ implies device-only inference.

Denote by $h_n$ the channel gain when the WD offloads the input data $I_n$ at the model splitting point $n$.  The noise at the receiver is additive white Gaussian noise (AWGN) with zero mean and variance $\sigma^2$. We assume that the transmit power of the WD is fixed as $P$. Suppose that the instantaneous signal-to-noise ratio (SNR) $\gamma_n=\frac{Ph_n}{\sigma^2}$ at model splitting point $n$ is random with probability density function (PDF) $f_n(\cdot)$ and cumulative distribution function (CDF) $F_n(\cdot)$.  We assume that the $\gamma_n$ is independent across different model splitting points. That is, the channel coherence time is comparable with the local computing time of one DNN layer.

Accordingly, the data transmission rate from the WD to the edge server at model splitting point $n$ is
\begin{align}
R_n(\gamma_n)=W\log_2(1+\gamma_n),
\end{align}
where $W$ is the fixed bandwidth allocated to the WD.
Then, the offloading transmission time at model splitting point $n$ is
\begin{align}
\tau_n^u(\gamma_n)=\frac{I_n}{R_n(\gamma_n)},
\end{align}
and the corresponding energy consumption is
\begin{align}
e_n^u(\gamma_n)=P\frac{I_n}{R_n(\gamma_n)}.
\end{align}

As for the on-device inference, we denote by $f_l$ the local CPU frequency for computing the subtasks. Then, the local execution time for subtask $i$ is
\begin{align}
\tau_i^l=\frac{L_i}{f_l},
\end{align}
and the corresponding energy consumption is
\begin{align}
e_i^l=\kappa L_if_l^2,
\end{align}
where $\kappa$ is the effective switched capacitance parameter depending on the chip architecture.

As for the edge inference, the edge computing time of subtask $i$ is
\begin{align}
\tau_i^c=\frac{L_i}{f_c},
\end{align}
where $f_c$ is the CPU frequency of the edge server.

\subsection{Problem Formulation}

In this subsection, we first formulate the optimal model splitting problem as an optimal stopping problem with finite horizon $M+1$. Then, we formulate the joint model placement and splitting optimization problem in \eqref{P1}.

\subsubsection{Optimal Stopping Problem With Given Model Placement Decision}
Suppose that the first $M$ layers of the DNN are placed at the WD. That is, the WD has to offload the intermediate features for edge inference no later than stage $M+1$.  Let $\bm{\gamma}=\{\gamma_1,...,\gamma_{M+1}\}$ denote a sequence of random variables representing the uplink SNR. Suppose that the WD can only observe the past and current SNRs, but not the future ones.  At stage $n$, having observed $\gamma_{n}$, the WD needs to decide whether to continue computing the $n$-th layer locally or stop local computing and offload the $n$-th layer's input data (i.e., the $(n-1)$-th layer's output data) to the edge server.

If the WD decides to stop at stage $n$ (i.e., split the model at point $n$), then the total energy-time cost (ETC), defined as the weighted sum of the WD's energy consumption and time on inference, is given by
\begin{align}
\eta_{n}(\gamma_n)&=\beta_t(\sum_{i=0}^{n-1}\tau_i^l+\sum_{i=n}^{N+1}\tau_i^c+\tau_n^u(\gamma_n))+\beta_e(\sum_{i=0}^{n-1}e_i^l+e_n^u(\gamma_n))\nonumber\\
&=\omega_n+(\beta_tI_{n}+\beta_ePI_{n})\frac{1}{R_n(\gamma_n)},
\end{align}
where
\begin{align}
\omega_n=\beta_t(\sum_{i=0}^{n-1}\frac{L_i}{f_l}+\sum_{i=n}^{N+1}\frac{L_i}{f_c})+\beta_e(\sum_{i=0}^{n-1}\kappa L_if_l^2).
\end{align}
Notice that $\omega_n$ increases in $n$ due to the more powerful computation capacity at the edge server (i.e., $f_l<f_c$). Besides, $\beta_t$ and $\beta_e$ denote the weights of total inference time and energy consumption, respectively. Note that the ETC $\eta_{n}(\gamma_n)$ is deterministically known to the WD since it observes the SNR $\gamma_n$ at stage $n$.

The WD decides to continue local computing at stage $n$ only when doing so incurs a lower ETC than stopping at stage $n$. Note that the ETC of stopping at a future stage is random to the WD, as it does not know the channel SNR in the future.

A stopping rule determines the model splitting point $S(M,\bm{\gamma})\in\{1,2,...,M+1\}$ based on the number of downloaded layers $M$ and the random SNR observations $\bm{\gamma}$.
Note that $S(M,\bm{\gamma})$ is random, as it is a function of random variables $\bm{\gamma}$. Different realizations of observations may lead to different stopping decisions. Given the model placement decision $M$, our purpose is to find the optimal model splitting strategy (i.e., optimal stopping rule) $S^*(M,\bm{\gamma})$ to minimize the expected inference ETC, i.e., $E_{\bm{\gamma}}[\eta_{S^*(M,\bm{\gamma})}(\gamma_{S^*(M,\bm{\gamma})})]$. In optimal stopping theory, this problem is a stopping rule problem with a finite horizon \cite{stopping}.

\subsubsection{Joint Optimization of Model Placement and Model Splitting Strategy}

Our goal is to find the optimal model splitting strategy $S^*(M,\bm{\gamma})$ and select the optimal number of downloaded layers $M, M\in\{0,1,...,N\}$, with the objective to minimize the overall expected cost, defined as the expected ETC of the WD plus the weighted average model downloading time cost, i.e.,
\begin{align}\label{Z}
Z(M,S^*(M,\bm{\gamma}))=\beta_t\psi(M)+E_{\bm{\gamma}}[\eta_{S^*(M,\bm{\gamma})}(\gamma_{S^*(M,\bm{\gamma})})].
\end{align}
Mathematically, the optimization problem is formulated as
\begin{eqnarray}\label{P1}
\mbox{(P1)}~~\min_{S^*(M,\bm{\gamma}),M}&&Z(M,S^*(M,\bm{\gamma})),\nonumber\\
{\rm s.t.}&& M\in\{0,1,...,N\},\nonumber\\
&& S^*(M,\bm{\gamma})\in\{1,2,...,M+1\}.
\end{eqnarray}
%

%In order to address the Problem (P1), we first assume fixed model placement decision $M$ and investigate the optimal model splitting strategy (i.e., the optimal stopping rule) $S^*(M,\bm{\gamma})$ in the following Section III.

\section{Optimal Model Placement and Online Model Splitting Strategy}

In this section, we first assume a fixed model placement decision $M$ and investigate the optimal model splitting strategy $S^*(M,\bm{\gamma})$. Then, based on the analysis of the expected ETC achieved by the optimal model splitting strategy, we propose the optimal model placement algorithm.

\subsection{Backward Induction}
Given a finite horizon $M+1$, we solve the optimal stopping problem by backward induction. Notice that when $M=0$, the WD has no choice but to offload the data $I_1$ to the edge server at the first stage. In the following, we consider the case where the number of downloaded layers $M\geq 1$.

With backward induction, we first find the optimal model splitting strategy at stage $M$. Then, we find the optimal splitting strategy at stage $M-1$, taking the decision at stage $M$ as given. The process continues backward until the first stage. Let $V_n^{(M+1)}$ denote the minimum expected inference ETC for splitting the model starting from stage $n, n\leq M$, given the current observation $\gamma_n$. That is,
\begin{align}\label{eq13}
V_n^{(M+1)}&=\min\left\{\eta_{n}(\gamma_n),E(V_{n+1}^{(M+1)})\right\}\nonumber\\
&=\left\{
    \begin{array}{ll}
      \eta_{n}(\gamma_n), & \gamma_n>\hat{\gamma}_{n}(M); \\
      E(V_{n+1}^{(M+1)}), & \gamma_n<\hat{\gamma}_{n}(M),
    \end{array}
  \right.
\end{align}
where
\begin{align}\label{prop3}
&\hat{\gamma}_{n}(M)=2^{\frac{(\beta_tI_{n}+\beta_ePI_{n})}{W\left[E(V_{n+1}^{(M+1)})-\omega_n\right]}}-1.
%&\phi_{n,\tilde{N}}=\left[E(V_{n+1}^{(\tilde{N}+1)})-\omega_n\right]/(\beta_tI_{n}+\beta_ePI_{n}).
\end{align}

At the last stage $M+1$, we have
\begin{small}
\begin{align}
E(V_{M+1}^{(M+1)})&=\int_{0}^{\infty}\left[\beta_t(\sum_{i=0}^{M}\tau_i^l+\sum_{i=M+1}^{N+1}\tau_i^c+\tau_{M+1}^u(\gamma_{M+1}))+\beta_e(\sum_{i=0}^{M}e_i^l+e_{M}^u(\gamma_{M+1}))\right]f_{M+1}(\gamma_{M+1})d\gamma_{M+1}\nonumber\\
&=\omega_{M+1}+(\beta_tI_{M+1}+\beta_ePI_{M+1})\int_{0}^{\infty}\frac{1}{R_{M+1}(\gamma_{M+1})}f_{M+1}(\gamma_{M+1})d\gamma_{M+1}.
\end{align}
\end{small}
Inductively, at stage $n$,
\begin{small}
\begin{align}\label{eq16}
E(V_{n}^{(M+1)})&=E\left[\min\left\{\eta_{n}(\gamma_n),E(V_{n+1}^{(M+1)})\right\}\right]\nonumber\\
&=\int_{\hat{\gamma}_{n}(M)}^{\infty}\eta_{n}(\gamma_n)f_n(\gamma_n)d\gamma_n+\int_{0}^{\hat{\gamma}_{n}(M)}E(V_{n+1}^{(M+1)})f_n(\gamma_n)d\gamma_n\nonumber\\
&=\omega_n(1-F(\hat{\gamma}_{n}(M)))+(\beta_tI_{n}+\beta_ePI_{n})\int_{\hat{\gamma}_{n}(M)}^{\infty}\frac{1}{R_n(\gamma_n)}f_n(\gamma_n)d\gamma_n+E(V_{n+1}^{(M+1)})F(\hat{\gamma}_{n}(M)).
\end{align}
\end{small}
%
%=&\psi_{n,\tilde{N}}+\int_{\hat{\gamma}_{n,\tilde{N}}}^{\infty}\left[\beta_t\sum_{i=n}^{\tilde{N}}\frac{L_i}{f_c}+(\beta_tI_{n}+\beta_ePI_{n})\frac{1}{R_{n}}\right]f(\gamma_n)d\gamma_n\nonumber\\
%&+\int_{0}^{\hat{\gamma}_{n,\tilde{N}}}\left[E(V_{n+1}^{(\tilde{N}+1)})-\psi_{n,\tilde{N}}\right]f(\gamma_n)d\gamma_n,
%where $\psi_{n,\tilde{N}}=\beta_t(\sum_{i=1}^{n-1}\frac{L_i}{f_l}+\sum_{i=\tilde{N}+1}^{N}\frac{L_i}{f_c})+\beta_t\sum_{i=1}^{\tilde{N}}\tau_i^m+\beta_e(\sum_{i=1}^{n-1}\kappa L_if_l^2)$.
Accordingly, we can calculate $E(V_{n}^{(M+1)})$ and $\hat{\gamma}_{n}(M)$ in \eqref{prop3} for all $n\leq M$. According to \eqref{eq13}, it is optimal to stop at stage $n$ if the observed SNR $\gamma_n>\hat{\gamma}_{n}(M)$, and to continue otherwise. In other words, the optimal stopping rule $S^*(M,\bm{\gamma})$ is a threshold-based policy determined by $\hat{\bm{\gamma}}(M)$, where $\hat{\bm{\gamma}}(M)=\{\hat{\gamma}_{1}(M),...,\hat{\gamma}_{M}(M)\}$. We therefore have the following Proposition 3.1.

\emph{\textbf{Proposition 3.1:}} Given the number of downloaded layers $M$, the optimal stopping rule $S^*(M,\bm{\gamma})$ is given by
\begin{align}
S^*(M,\bm{\gamma})=\left\{
         \begin{array}{ll}
           \min\{\Psi\}, & \Psi\neq\emptyset \\
           M+1, & \hbox{otherwise,}
         \end{array}
       \right.
\end{align}
where
\begin{align}
\Psi=\{n|1\leq n\leq M, \gamma_{n}\geq\hat{\gamma}_n(M)\},
\end{align}
and $\hat{\gamma}_n(M)$ is defined in \eqref{prop3}.

According to Proposition 3.1, when $\Psi\neq\emptyset$, our proposed online model splitting strategy is to split the DNN model at the first stage that the observed SNR $\gamma_{n}$ is larger than or equal to the corresponding threshold $\hat{\gamma}_n(M)$.
In addition, $S^*(M,\bm{\gamma})=M+1$ when $\Psi=\emptyset$.
%That is, if the number of downloaded layers $M=0$, the optimal stopping rule calls for stopping at stage 1. Besides, if the WD decides to continue from stage 1 to stage $M$ (i.e., $\gamma_{n}\leq\hat{\gamma}_n(M), \forall n\in\{1,...,M\}$), the optimal stopping rule calls for stopping at the last stage $M+1$.

\subsection{Expected Inference ETC Performance}

In this subsection, we analyze the expected ETC $E_{\bm{\gamma}}[\eta_{S^*(M,\bm{\gamma})}(\gamma_{S^*(M,\bm{\gamma})})]$ achieved by the optimal model spitting. If $M=0$, the expected ETC $E_{\bm{\gamma}}[\eta_{S^*(M,\bm{\gamma})}(\gamma_{S^*(M,\bm{\gamma})})]=E(V^{(1)}_1)$. In the following, we consider the case where $M\geq 1$.
Specifically, we denote the probability of stopping at stage $n$ as $Pr\{S^*(M,\bm{\gamma})=n\}$. Then, we have
\begin{align}\label{combine1}
Pr\{S^*(M,\bm{\gamma})=n\}=\left\{
       \begin{array}{ll}
         1-F_1(\hat{\gamma}_{1}(M)), & n=1; \\
         \left(\prod_{j=1}^{n-1}F_j(\hat{\gamma}_{j}(M))\right)\left(1-F_n(\hat{\gamma}_{n}(M))\right), & 1<n<M+1; \\
         \prod_{j=1}^{M}F_j(\hat{\gamma}_{j}(M)), & n=M+1.
       \end{array}
     \right.
\end{align}
Besides, the expected ETC $\eta_n(\gamma_n)$ when $S^*(M,\bm{\gamma})=n$ is calculated as
\begin{align}\nonumber
E_{\bm{\gamma}}[\eta_{S^*(M,\bm{\gamma})}|S^*(M,\bm{\gamma})=n]=\left\{
                  \begin{array}{ll}
                    \omega_n+E_{\gamma_n}\left[(\beta_tI_{n}+\beta_ePI_{n})\frac{1}{R_n(\gamma_n)}|\gamma_n>\hat{\gamma}_{n}(M)\right], & n\leq M; \\
                    \omega_{n}+E_{\gamma_n}\left[(\beta_tI_{n}+\beta_ePI_{n})\frac{1}{R_{n}(\gamma_{n})}\right], & n=M+1.
                  \end{array}
                \right.
\end{align}
That is,
\begin{align}\label{combine2}
E_{\bm{\gamma}}[\eta_{S^*(M,\bm{\gamma})}|S^*(M,\bm{\gamma})=n]=\left\{
                  \begin{array}{ll}
                    \omega_n+(\beta_tI_{n}+\beta_ePI_{n})\frac{\int_{\hat{\gamma}_{n}(M)}^{\infty}\frac{1}{R_n(\gamma_n)}f_n(\gamma_n)d\gamma_n}{1-F_n(\hat{\gamma}_{n}(M))}, & n\leq M; \\
                    \omega_{n}+(\beta_tI_{n}+\beta_ePI_{n})\int_{0}^{\infty}\frac{1}{R_{n}(\gamma_{n})}f_n(\gamma_{n})d\gamma_{n}, & n=M+1.
                  \end{array}
                \right.
\end{align}
%
%%
%\begin{align}
%E_{n,\tilde{N}}&=\omega_n+\beta_t\sum_{i=1}^{\tilde{N}}\tau_i^m+E\left((\beta_tI_{n}+\beta_ePI_{n})\frac{1}{R_n(\gamma_n)}|\gamma_n>\hat{\gamma}_{n,\tilde{N}}\right)\nonumber\\
%&=\omega_n+\beta_t\sum_{i=1}^{\tilde{N}}\tau_i^m+(\beta_tI_{n}+\beta_ePI_{n})\frac{\int_{\hat{\gamma}_{n,\tilde{N}}}^{\infty}\frac{1}{R_n(\gamma_n)}f(\gamma_n)d\gamma_n}{1-F(\hat{\gamma}_{n,\tilde{N}})}.
%\end{align}
%%
Therefore, the expected ETC given that the first $M$ layers are downloaded to the WD is
\begin{align}\label{combine}
E_{\bm{\gamma}}[\eta_{S^*(M,\bm{\gamma})}(\gamma_{S^*(M,\bm{\gamma})})]=\sum_{n=1}^{M+1}Pr\{S^*(M,\bm{\gamma})=n\}E_{\bm{\gamma}}[\eta_{S^*(M,\bm{\gamma})}|S^*(M,\bm{\gamma})=n], M\geq 1.
\end{align}
By substituting \eqref{combine1} and \eqref{combine2} into \eqref{combine}, the expected ETC $E_{\bm{\gamma}}[\eta_{S^*(M,\bm{\gamma})}(\gamma_{S^*(M,\bm{\gamma})})]$, and hence the total expected cost $Z(M)$ in \eqref{Z}, can be expressed as a function of $M$.

\subsection{Optimal Model Placement}

In this subsection, we are ready to optimize the model placement decision based on the optimal stopping rule $S^*(M,\bm{\gamma})$ derived in Proposition 3.1 and the analysis of expected ETC in Section III B. Intuitively, one can enumerate all feasible $M\in\{0,1,...,N\}$ and select the optimal one that achieves the minimal expected cost $Z(M)$.

Notice that the decision threshold $\hat{\gamma}_n(M)$ for the optimal stopping rule involves a nested expectation of ETC in future stages (see \eqref{prop3} and \eqref{eq16}). For given $M$, it takes $O(M)$ complexity to compute $\hat{\gamma}_n(M)$ using backward induction. Then, by exhausting all feasible model placement decisions, the joint optimization of model placement and splitting for solving Problem (P1) incurs a polynomial computational time complexity $O(N^2)$.

\emph{\textbf{Remark 3.1:}} Notice that the optimal model placement decision and online model splitting strategy obtained by solving Problem (P1) remain the same as long as the DNN model structure, the model update frequency, and the statistic characteristics of wireless channels do not change. We claim that although solving Problem (P1) incurs $O(N^2)$ computational complexity, the online implementation for the model splitting follows a simple threshold-based policy with low complexity. Once the optimal model placement decision is determined, the corresponding thresholds $\hat{\bm{\gamma}}(M)$ for the optimal stopping rule are simultaneously obtained and stored at the WD. This facilitates the online model splitting for every DNN inference according to Proposition 3.1.

\section{Reduced-Complexity Algorithms}

The optimal model placement algorithm through exhaustive search results in $O(N^2)$ computational complexity due to the full process of backward induction when evaluating each feasible model placement decision. In this section, we are motivated to investigate linear-complexity, i.e., $O(N)$ algorithms based on a one-stage look-ahead stopping rule.
%based on which a reduced-complexity searching method is proposed to optimize $M$ in linear time complexity.

\subsection{One-Stage Look-Ahead Stopping Rule for Model Splitting}

First, we introduce the definition of one-stage look-ahead stopping rule.

\emph{\textbf{Definition 1 (one-stage look-ahead stopping rule):}} The one-stage look-ahead (1-sla) stopping rule is the one that stops if the cost for stopping at this stage is no more than the expected cost of continuing one stage and then stopping. Mathematically, the 1-sla stopping rule is described by the stopping time \cite{stopping}
\begin{align}
S_1=\min\{n\geq 1:\eta_n\leq E_{\gamma_{n+1}}[\eta_{n+1}|\gamma_1,...,\gamma_n]\}.
\end{align}

Note that the backward-induction based optimal stopping rule in Section III accounts for the ETC for all future stages until the last one. In contrast, the decision making at each stage under the 1-sla stopping rule only depends on the current observation and the expected performance of continuing for just one stage.
%Motivated by that, we propose the 1-sla stopping rule based model placement algorithm to reduce the computational complexity.

In the following, we first derive the 1-sla stopping rule for solving (P1) given the number of downloaded layers $M$.

\emph{\textbf{Proposition 4.1:}} Given the number of downloaded layers $M$, the 1-sla stopping rule $S_1(M,\bm{\gamma})$ is given by
\begin{align}\label{onesla}
S_1(M,\bm{\gamma})=\left\{
         \begin{array}{ll}
           \min\{\Omega\}, & \Omega\neq\emptyset \\
           M+1, & \hbox{otherwise,}
         \end{array}
       \right.
\end{align}
where
\begin{align}
\Omega=\{n|1\leq n\leq M, \gamma_{n}\geq\hat{\gamma}_n^{1-sla}\},
\end{align}
and $\hat{\gamma}_n^{1-sla}$ is given by
\begin{align}\label{shr_onesla}
\hat{\gamma}_n^{1-sla}=2^{\frac{1}{W}\frac{\beta_tI_{n}+\beta_ePI_{n}}{I_{n+1}(\beta_t+\beta_eP)\int_{0}^{\infty}\frac{1}{R_{n+1}}f_{n+1}(\gamma_{n+1})d\gamma_{n+1}+\beta_t(\frac{L_n}{f_l}-\frac{L_n}{f_c})+\beta_e\kappa L_n f^2_l}}-1.
\end{align}

\begin{proof}
According to Definition 1, the 1-sla stopping rule is
\begin{eqnarray}
S_1(M,\bm{\gamma})&=\min\bigg\{1\leq n\leq M: &\eta_{n}\leq E[V_{n+1}^{(n+1)}]\bigg\}\nonumber\\
&=\min\bigg\{1\leq n\leq M: &\eta_{n}\leq E_{\gamma_{n+1}}[\eta_{n+1}|\gamma_1,...,\gamma_n]\bigg\}\nonumber\\
&=\min\bigg\{1\leq n\leq M: &\omega_n+(\beta_tI_{n}+\beta_ePI_{n})\frac{1}{R_{n}}\leq \int_0^{\infty}\eta_{n+1}(\gamma_{n+1})f_{n+1}(\gamma_{n+1})d\gamma_{n+1}\bigg\}\nonumber\\
&=\min\bigg\{1\leq n\leq M:&\gamma_{n}\geq\hat{\gamma}_n^{1-sla} \bigg\},
\end{eqnarray}
where $\hat{\gamma}_n^{1-sla}$ is given in \eqref{shr_onesla}.

When $\Omega=\{n|1\leq n\leq M, \gamma_{n}\geq\hat{\gamma}_n^{1-sla}\}=\emptyset$, then $S_1(M,\bm{\gamma})=M+1$.
%That is, if the number of downloaded layers $M=0$, the 1-sla stopping rule calls for stopping at the first stage. Besides, if the WD decides to continue from stage 1 to stage $M$ (i.e., $\gamma_{n}\leq\hat{\gamma}_n^{1-sla}, \forall n\in\{1,...,M\}$), the 1-sla stopping rule calls for stopping at the last stage $M+1$.
\end{proof}

From Proposition 4.1, we have the following observations:
\begin{itemize}
  \item The decision made in each stage $n\in[1,M]$ in the 1-sla stopping rule only depends on the expectation of the unit transmission delay $\frac{1}{R_{n+1}}$ at the next stage $n+1$ and the model parameters in the $n$-th layer (i.e., the input and output data sizes of the $n$-th layer $I_{n},I_{n+1}$ and the computation workload $L_n$ of the $n$-th layer), regardless of $M$.
  \item Compared with the calculation of $\hat{\gamma}_n$ in the optimal stopping rule via backward induction, the decision threshold $\hat{\gamma}_n^{1-sla}$ in $S_1(M,\bm{\gamma})$ is much easier to calculate without the nested structure for the expected ETC in all future stages.
  \item The probability of stopping at stage $n$ decreases when the input data size $I_n$ of stage $n$ increases, the input data size of the next stage $I_{n+1}$ is smaller, or the difference between the local and edge execution costs for layer $n$ is lower.
\end{itemize}

According to Proposition 4.1, when $\Omega\neq\emptyset$, the 1-sla stopping rule based online model splitting scheme is to split the model at the first point where $\gamma_{n}\geq\hat{\gamma}_n^{1-sla}$ occurs.

The 1-sla stopping rule is not optimal in general. According to \cite{stopping}, the 1-sla rule is optimal in a finite horizon monotone stopping rule problem.  In the following Lemma 4.1 and Proposition 4.2, we show that the proposed 1-sla based model splitting strategy $S_1(M,\bm{\gamma})$ is optimal with certain probability.

\emph{\textbf{Lemma 4.1:}} Let $A_n$ denote the event $\{\eta_n\leq E[\eta_{n+1}|\gamma_1,...,\gamma_n]\}$, i.e., the 1-sla calls for stopping at stage $n$, for a given $n\in [1,M]$. Suppose that $A_n$ holds. Then, if $A_{n+1},...,A_{M}$ also hold, the 1-sla stopping rule $S_1(M,\bm{\gamma})$ is optimal, i.e., $S_1(M,\bm{\gamma})=S^*(M,\bm{\gamma})$.

\begin{proof}
Recall that the optimal stopping rule is
\begin{align}\nonumber
S^*(M,\bm{\gamma})=\min\{n\geq 1: \eta_n\leq E(V_{n+1}^{(M+1)}|\gamma_1,...,\gamma_n)\},
\end{align}
where $V_{M+2}^{(M+1)}=+\infty$, $V_{M+1}^{(M+1)}=\eta_{M+1}$, and by backward induction,
\begin{align}\nonumber
V_{n}^{(M+1)}=\min\{\eta_n,E(V_{n+1}^{(M+1)}|\gamma_1,...,\gamma_n)\},
\end{align}
for $n=1,...,M$.

Suppose that the 1-sla calls for stopping at stage $n$ and $A_{n+1},...,A_{M}$ also hold. First, for $A_{M}$, we have
\begin{align}\nonumber
\eta_{M}\leq E(\eta_{M+1}|\gamma_1,...,\gamma_{M})=E(V_{M+1}^{(M+1)}|\gamma_1,...,\gamma_{M}).
\end{align}
Hence,
\begin{align}\nonumber
V_{M}^{(M+1)}=\min\{\eta_M,E(V_{M+1}^{(M+1)}|\gamma_1,...,\gamma_M)\}=\eta_M.
\end{align}
Then, for $A_{M-1}$, we have
\begin{align}\nonumber
\eta_{M-1}\leq E(\eta_{M}|\gamma_1,...,\gamma_{M-1})=E(V_{M}^{(M+1)}|\gamma_1,...,\gamma_{M-1}).
\end{align}
Hence,
\begin{align}\nonumber
V_{M-1}^{(M+1)}=\min\{\eta_{M-1},E(V_{M}^{(M+1)}|\gamma_1,...,\gamma_{M-1})\}=\eta_{M-1}.
\end{align}
Similarly, for $n< k<M-1$, we have $V_{k}^{(M+1)}=\eta_{k}$. Finally, for $A_n$, we have
\begin{align}\nonumber
\eta_{n}\leq E(\eta_{n+1}|\gamma_1,...,\gamma_{n})=E(V_{n+1}^{(M+1)}|\gamma_1,...,\gamma_{n}).
\end{align}
Hence,
\begin{align}\nonumber
V_{n}^{(M+1)}=\min\{\eta_{n},E(V_{n+1}^{(M+1)}|\gamma_1,...,\gamma_{n})\}=\eta_{n}.
\end{align}
Therefore, the optimal stopping rule $S^*(M,\bm{\gamma})$ also calls for stopping at stage $n$.
\end{proof}

Based on the above lemma, we are ready to derive the probability that 1-sla stopping rule is optimal.

\emph{\textbf{Proposition 4.2:}} Given the number of downloaded layers $M$, the 1-sla stopping rule $S_1(M,\bm{\gamma})$ obtained by \eqref{onesla} is optimal with probability
\begin{align}\label{1-sla_opt}
Pr\{S_1(M,\bm{\gamma})=S^*(M,\bm{\gamma})\}=\sum_{n=1}^{M+1}\left[\prod_{j=1}^{n-1}F_j(\hat{\gamma}_j^{1-sla})\prod_{k=n}^{M}[1-F_k(\hat{\gamma}_k^{1-sla})]\right].
\end{align}

\begin{proof}
%To start with, we give the following definition and theorem \cite{stopping}.
%
%\emph{Definition 1:} Let $A_n$ denote the event $\{\eta_n\leq E[\eta_{n+1}|\gamma_1,...,\gamma_n]\}$. We say that the stopping problem is monotone if
%%
%\begin{align}
%A_0\subset A_1\subset A_2\subset ...
%\end{align}
%%
%
%For the monotone 1-sla stopping rule problem, if the 1-sla stopping rule calls for stopping at stage $n$, then it will call for stopping at all future stages no matter what the future observations turn out to be.
%
%\emph{Theorem 1:} In a finite horizon monotone stopping rule problem, the 1-sla rule is optimal.
If the 1-sla stopping rule $S_1(M,\bm{\gamma})$ obtained by \eqref{onesla} calls for stopping at stage $n$, then it will also call for stopping at all future stages with probability
\begin{align}
Pr\{\mbox{Stage $n$ is reached and}~A_n,A_{n+1}, ...,A_{M} ~\mbox{hold}\}=\prod_{j=1}^{n-1}F_j(\hat{\gamma}_j^{1-sla})\prod_{k=n}^{M}[1-F_k(\hat{\gamma}_k^{1-sla})].
\end{align}
Then, according to Lemma 4.1, we can obtain the probability $Pr\{S_1(M,\bm{\gamma})=S^*(M,\bm{\gamma})\}$ as shown in \eqref{1-sla_opt}.
\end{proof}

\emph{\textbf{Corollary 4.1:}} When the number of downloaded layers $M=1$, the 1-sla stopping rule is optimal, i.e., $S_1(M,\bm{\gamma})=S^*(M,\bm{\gamma})$.

\begin{proof}
According to Proposition 4.2, when $M=1$, we have
\begin{align}
Pr\{S_1(1,\bm{\gamma})=S^*(1,\bm{\gamma})\}=[1-F_1(\hat{\gamma}_1^{1-sla})]+F_1(\hat{\gamma}_1^{1-sla})=1.
\end{align}
Hence, when $M=1$, the 1-sla stopping rule is optimal.
\end{proof}

\subsection{1-sla Stopping Rule Based Model Placement}

Following the similar technique in Section III B, we can analyze the expected ETC performance $E_{\bm{\gamma}}[\eta_{S_1(M,\bm{\gamma})}(\gamma_{S_1(M,\bm{\gamma})})]$, and hence the overall expected cost $Z(M)$, as a function of model placement $M$ under the 1-sla stopping rule.  Then, we can enumerate all feasible $M\in\{0,1,...,N\}$ to find $M^*$ that yields the minimal expected cost $Z(M)$.

It takes $O(N)$ complexity to compute all possible $\{\hat{\gamma}_n^{1-sla}, \forall n\in [1,N]\}$ in $S_1(M,\bm{\gamma})$ according to \eqref{shr_onesla}. Besides, the complexity of enumerating all feasible model placement decisions is $O(N)$.  Notice that the decision threshold $\hat{\gamma}_n^{1-sla}$ under the 1-sla stopping rule is independent of the model placement $M$.  Therefore, the overall computational complexity for solving Problem (P1) reduces to linear complexity $O(N)$ under the 1-sla stopping rule based algorithm.

\subsection{Case Study: Fully-Connected Multilayer Perceptron}

To obtain more engineering insights, we consider fully-connected multilayer perceptron (MLP) networks in this subsection. We will show that under certain assumptions, the closed-form expressions of the optimal model placement $M^*$ can be derived when the model splitting decision is based on the 1-sla stopping rule.

Fully-connected MLP \cite{mlp} is a class of feedforward artificial neural network, which consists of an input layer, multiple hidden layers and an output layer. Except for the nodes of the input layer, each node of the other layers is a neuron that uses a nonlinear activation function.

\begin{figure}[htb]
\begin{centering}
\includegraphics[scale=0.3]{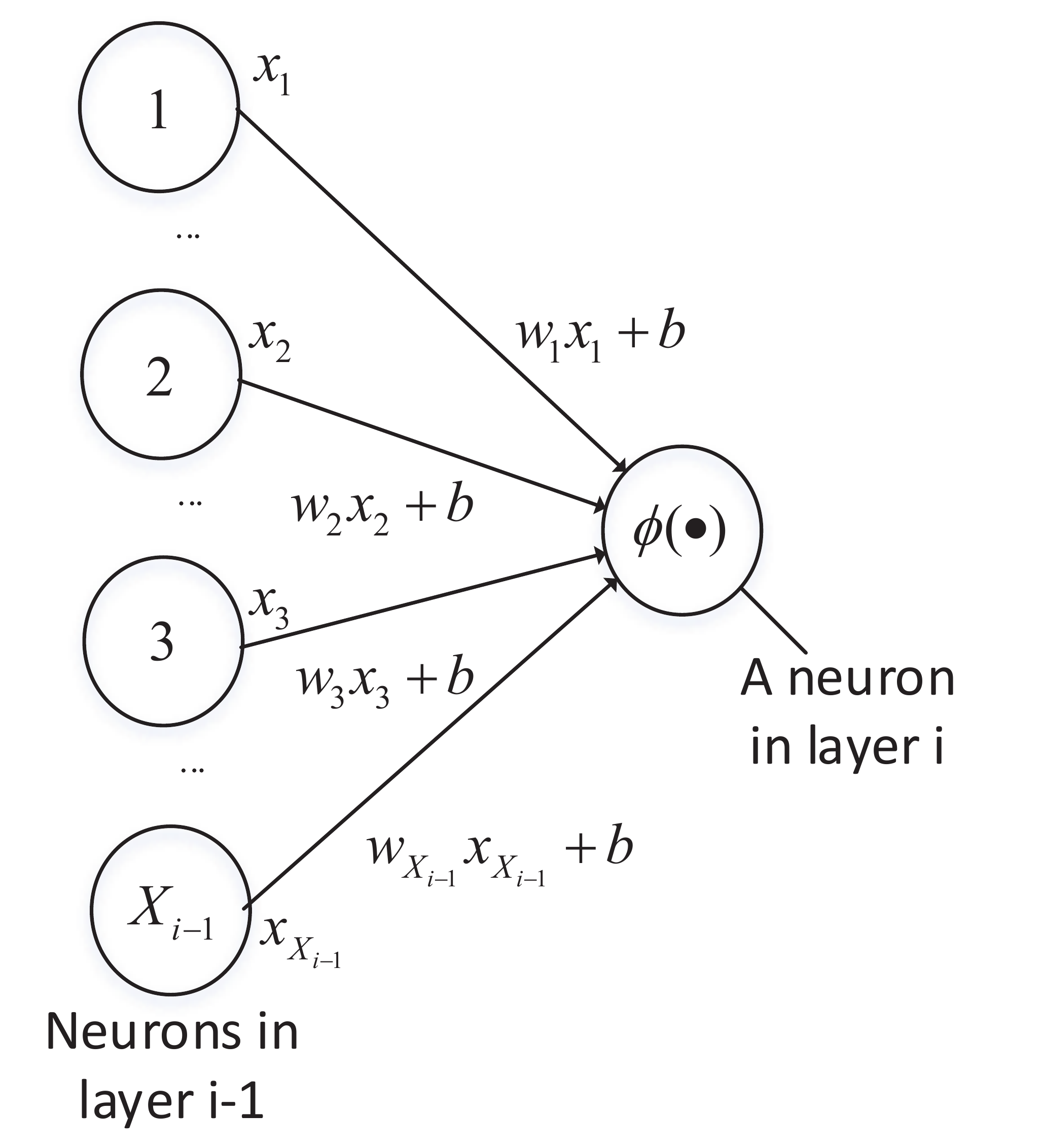}
\vspace{-0.1cm}
 \caption{An illustration of the floating-point multiply-add calculations in one neuron of layer $i$, where $\{w_1,w_2,...\}$ and $b$ are the parameters corresponding to that neuron in layer $i$ and $\phi(\cdot)$ is the nonlinear activation function.}\label{MLP}
\end{centering}
\vspace{-0.1cm}
\end{figure}

Suppose that layer $i$ has $X_i$ neurons with input data size
\begin{align}\label{def_lamb}
I_i=\lambda X_{i-1},
\end{align}
where $\lambda$ is the number of bytes to represent the original output value of each neuron. In Fig. \ref{MLP}, we illustrate the floating-point multiply-add calculations in one neuron of layer $i$. Each neuron in layer $i$ first performs $X_{i-1}$ multiply-add operations on the input data received from the previous layer using the weights (e.g., $\{w_1,w_2,...,w_{X_{i-1}}\}$ in Fig. 4) and biases (e.g., $b$ in Fig. 4) defined by the model parameters. Then, the result is further processed by a nonlinear activation function (e.g., $\phi(\cdot)$ in Fig. 4). It can be seen that the number of model parameters required for the inference of layer $i$ and the total computation workload of layer $i$ depend on the number of neurons in layer $i-1$ and $i$.

From the above, we see that the computation workload of each layer is proportional to $X_{i-1}X_i$, i.e.,
\begin{align}
L_i=\alpha X_{i-1}X_i,
\end{align}
where $\alpha$ is the number of CPU cycles required to execute one floating-point multiply-add operation. Likewise, the number of model parameters needed by layer $i$, including the weights and biases, is $(X_{i-1}+1)X_i$. Hence, the parameter downloading time for layer $i$ is
\begin{align}\label{def_mu}
\tau_{i}^m=\frac{\mu (X_{i-1}+1)X_i}{R^{d}},
\end{align}
where $\mu$ is the number of bytes to represent each model parameter and $R^d$ is the downlink transmission data rate. In this paper, we assume that the transmit power of the BS is fixed and the downlink channel gain follows the free-space path loss model.
%and the number of neurons between the neighbor layers has the relation $X_{i+1}=\nu X_i$.

%Before the analysis, we first state the following assumptions.
%
%\emph{\textbf{Assumption 1:}} In this subsection, we assume that the computational workload of each layer is proportional to its input data size. Let $\alpha$ denote the number of CPU cycles required to compute one bit task. Accordingly, the total number of CPU cycles required to execute layer $i$ is $L_i=\alpha I_i$.

%\emph{\textbf{Assumption 1:}} The downloading time of layer $i$ is proportional to the input data size of layer $i$, i.e., $\tau_{i}^m=\mu I_{i}$.

In the following, we assume that the number of neurons in each layer remains the same (i.e., $X_{i-1}=X_i, \forall i$).
For the simplicity of analysis, we assume that the SNR $\gamma_n$ in each stage is identically distributed, i.e., $f_1(\cdot)=f_2(\cdot)=...=f(\cdot)$ and $F_1(\cdot)=F_2(\cdot)=...=F(\cdot)$.

\emph{\textbf{Lemma 4.2:}} For a fully-connected MLP with $X_{i-1}=X_i=X, \forall i$, the 1-sla stopping threshold $\hat{\gamma}_n^{1-sla}$ in \eqref{shr_onesla} is a constant, i.e., $\hat{\gamma}_1^{1-sla}=...=\hat{\gamma}_N^{1-sla}=\delta(X)$, where
\begin{align}
\delta(X)=2^{\frac{1}{W}\frac{(\beta_t+\beta_eP)\lambda}{\lambda(\beta_t+\beta_eP)\int_{0}^{\infty}\frac{1}{R_{n+1}}f(\gamma_{n+1})d\gamma_{n+1}+[\beta_t(\frac{1}{f_l}-\frac{1}{f_c})+\beta_e\kappa f^2_l]\alpha X}}-1.
\end{align}

Lemma 4.2 indicates that for a fully-connected MLP with fixed number of neurons in each layer, the 1-sla stopping rule is to compare the observed SNR $\gamma_n$ with a constant $\delta(X)$.

To derive the model placement solution, the following lemma studies the relation between the expected ETC $E_{\bm{\gamma}}[\eta_{S_1(M,\bm{\gamma})}(\gamma_{S_1(M,\bm{\gamma})})]$ and the number of downloaded layers $M$ under the 1-sla stopping rule.

\emph{\textbf{Lemma 4.3:}} With the increase of $M$, the expected inference ETC $E_{\bm{\gamma}}[\eta_{S_1(M,\bm{\gamma})}(\gamma_{S_1(M,\bm{\gamma})})]$  under the 1-sla rule decreases. That is,
\begin{align}\label{bigtheta}
\Theta^{1-sla}(M)=E_{\bm{\gamma}}[\eta_{S_1(M,\bm{\gamma})}(\gamma_{S_1(M,\bm{\gamma})})]-E_{\bm{\gamma}}[\eta_{S_1(M-1,\bm{\gamma})}(\gamma_{S_1(M-1,\bm{\gamma})})]<0, \forall M\in [1,N].
\end{align}

\begin{proof}
The difference between the expected ETC when downloading $M$ and $M-1$ layers is calculated as
\begin{small}
\begin{eqnarray}
\Theta^{1-sla}(M)&=&E_{\bm{\gamma}}[\eta_{S_1(M,\bm{\gamma})}(\gamma_{S_1(M,\bm{\gamma})})]-E_{\bm{\gamma}}[\eta_{S_1(M-1,\bm{\gamma})}(\gamma_{S_1(M-1,\bm{\gamma})})]\nonumber\\
&=&\left(\prod_{j=1}^{M-1}F(\hat{\gamma}_{j}^{1-sla})\right)\bigg[(1-F(\hat{\gamma}_{M}^{1-sla}))\left(\omega_{M}+(\beta_tI_{M}+\beta_ePI_{M})\frac{\int_{\hat{\gamma}_{M}^{1-sla}}^{\infty}\frac{1}{R_{M}(\gamma_{M})}f(\gamma_{M})d\gamma_{M}}{1-F(\hat{\gamma}_{M}^{1-sla})}\right)\nonumber\\
&&+F(\hat{\gamma}_{M}^{1-sla})\left(\omega_{M+1}+(\beta_tI_{M+1}+\beta_ePI_{M+1})\int_{0}^{\infty}\frac{1}{R_{M+1}(\gamma_{M+1})}f(\gamma_{M+1})d\gamma_{M+1}\right)\nonumber\\
&&-\left(\omega_{M}+(\beta_tI_{M}+\beta_ePI_{M})\int_{0}^{\infty}\frac{1}{R_{M}(\gamma_{M})}f(\gamma_{M})d\gamma_{M}\right)\bigg]\nonumber\\
&=&\left(\prod_{j=1}^{M}F(\hat{\gamma}_{j}^{1-sla})\right)\bigg[\omega_{M+1}+(\beta_tI_{M+1}+\beta_ePI_{M+1})\int_{0}^{\infty}\frac{1}{R_{M+1}(\gamma_{M+1})}f(\gamma_{M+1})d\gamma_{M+1}\nonumber\\
&&-\left(\omega_{M}+(\beta_tI_{M}+\beta_ePI_{M})\frac{\int_{0}^{\hat{\gamma}_{M}^{1-sla}}\frac{1}{R_{M}(\gamma_{M})}f(\gamma_{M})d\gamma_{M}}{F(\hat{\gamma}_{M}^{1-sla})}\right)\bigg].
\end{eqnarray}
\end{small}
According to \eqref{onesla}, $\forall \gamma_{M}<\hat{\gamma}_{M}^{1-sla}$, we have
\begin{small}
\begin{align}\nonumber
\omega_{M}+(\beta_tI_{M}+\beta_ePI_{M})\frac{1}{R_{M}(\gamma_{M})}>\omega_{M+1}+(\beta_tI_{M+1}+\beta_ePI_{M+1})\int_{0}^{\infty}\frac{1}{R_{M+1}(\gamma_{M+1})}f(\gamma_{M+1})d\gamma_{M+1}.
\end{align}
\end{small}
Therefore, we have $\Theta^{1-sla}(M)<0, \forall M\in [1,N]$.
%Similarly, when $M=N$, we have
%%
%\begin{eqnarray}\nonumber
%\Theta^{1-sla}(N)=\left(\prod_{j=1}^{N}F(\hat{\gamma}_{j}^{1-sla})\right)\bigg[\omega_{N+1}-\left(\omega_{N}+(\beta_tI_{N}+\beta_ePI_{N})\frac{\int_{0}^{\hat{\gamma}_{N}^{1-sla}}\frac{1}{R_{N}(\gamma_{N})}f(\gamma_{N})d\gamma_{N}}{F(\hat{\gamma}_{N}^{1-sla})}\right)\bigg]<0.
%\end{eqnarray}
%%
%Hence, we have $\Theta^{1-sla}(M)<0, \forall 1\leq M\leq N$.
\end{proof}

From Lemma 4.3, we observe that when the edge server places more layers to the WD (i.e., larger $M$), the expected ETC decreases. Nevertheless, according to \eqref{download}, a larger downloading time cost $\psi(M)$ occurs when $M$ increases.

\emph{\textbf{Corollary 4.2:}} When the model update frequency $K$ is sufficiently large, the optimal $M^*=N$.

\begin{proof}
When $K$ is sufficiently large, the average model downloading time cost $\psi(M)=\frac{\sum_{i=1}^{M}\tau_i^m}{K}$ diminishes. Then, according to Lemma 4.3, we have $M^*=N$.
\end{proof}

Based on Lemma 4.3, we are ready to derive the optimal model placement solution in the following proposition.

%%%%%%%%%%%%
%\emph{\textbf{Lemma 3:}} For the fully-connected MLP with $X_{i-1}=X_i=X, \forall i$, $|\Theta^{1-sla}(M)|$ decreases with $M$ for the 1-sla rule.
%
%
%
%
%\begin{proof}
%%If the deep neural network structures satisfy $I_{n+1}=\nu I_n, \forall n$, $\hat{\gamma}_n^{1-sla}$ is a constant
%Let $\Lambda(M)=\Theta^{1-sla}(M)-\Theta^{1-sla}(M-1)$, where $2\leq M\leq N$. Then, we have
%%For $M=N$, we have
%%%
%%\begin{eqnarray}\label{lemma2}
%%\Lambda(N)&=&\Theta^{1-sla}(N)-\Theta^{1-sla}(N-1)\nonumber\\
%%&=&\left(\prod_{j=1}^{N-1}F(\delta(\nu))\right)I_{N-1}\bigg\{F(\hat{\gamma}_{N}^{1-sla})\nu\bigg[\alpha\beta_t(\frac{1}{f_l}-\frac{1}{f^c})+\alpha\beta_e\kappa f_l^2-(\beta_t+\beta_eP)\nonumber\\
%%&&\frac{\int_{0}^{\hat{\gamma}_{N}^{1-sla}}\frac{1}{R_{N}(\gamma_{N})}f(\gamma_{N})d\gamma_{N}}{F(\hat{\gamma}_{N}^{1-sla})})\bigg]-\bigg[\alpha\beta_t(\frac{1}{f_l}-\frac{1}{f^c})+\alpha\beta_e\kappa f_l^2+(\beta_t+\beta_eP)\nonumber\\
%%&&(\nu\int_{0}^{\infty}\frac{1}{R_{N}(\gamma_{N})}f(\gamma_{N})d\gamma_{N}-\frac{\int_{0}^{\delta(\nu)}\frac{1}{R_{N-1}(\gamma_{N-1})}f(\gamma_{N-1})d\gamma_{N-1}}{F(\delta(\nu))})\bigg]\bigg\}.
%%\end{eqnarray}
%%%
%%
%\begin{align}
%\Lambda(M)=\left(\prod_{j=1}^{M-1}F(\delta(X))\right)Xg(\delta(X))\left[ F(\delta(X)) -1\right].
%\end{align}
%%
%According to Lemma 2, we have $g(\delta(\nu))<0$. Therefore, we have $\Lambda(M)>0$. Since $\Theta^{1-sla}(M)<0, \forall M$, $|\Theta^{1-sla}(M)|$ decreases with $M$.
%\end{proof}

\emph{\textbf{Proposition 4.3:}} If the fully-connected MLP satisfies $X_{i-1}=X_i=X, \forall i$ and the model splitting is based on the 1-sla stopping rule, then the optimal number of downloaded layers is given by
\begin{align}\label{opt_layer}
M^*=\left\{
              \begin{array}{ll}
                N, & [F(\delta(X))]^{N}g(\delta(X))+\frac{\mu (X+1)}{KR^d}<0;\\
                0, & F(\delta(X))g(\delta(X))+\frac{\mu (X+1)}{KR^d}>0; \\
                \left\langle\log_{F(\delta(X))}\left(\frac{\mu (X+1)}{-KR^dg(\delta(X))}\right)\right\rangle, & \hbox{otherwise,}
              \end{array}
            \right.
\end{align}
where $\left\langle\cdot\right\rangle$ is the rounding function and $g(\delta(X))$ is given in \eqref{eq37}.

\begin{proof}
For the fully-connected MLP with $X_{i-1}=X_i=X, \forall i$, we first rewrite $\Theta^{1-sla}(M)$ in \eqref{bigtheta} as
\begin{align}\label{simplify_gv}
\Theta^{1-sla}(M)=X[F(\delta(X))]^{M}g(\delta(X)), 1\leq M\leq N,
\end{align}
where
\begin{eqnarray}\label{eq37}
g(\delta(X))&=&\alpha X\beta_t(\frac{1}{f_l}-\frac{1}{f^c})+\alpha X\beta_e\kappa f_l^2+(\beta_t+\beta_eP)\lambda (\int_{0}^{\infty}\frac{1}{R_{M+1}(\gamma_{M+1})}f(\gamma_{M+1})d\gamma_{M+1}\nonumber\\
&&-\frac{\int_{0}^{\delta(X)}\frac{1}{R_{M}(\gamma_{M})}f(\gamma_{M})d\gamma_{M}}{F(\delta(X))})\nonumber\\
&=&(\beta_t+\beta_eP)\lambda\left(\frac{1}{R_M(\delta(X))}-\frac{\int_{0}^{\delta(X)}\frac{1}{R_{M}(\gamma_{M})}f(\gamma_{M})d\gamma_{M}}{F(\delta(X))}\right).
\end{eqnarray}

Then, we have
\begin{align}
\Delta^{1-sla}(M)&=Z(M)-Z(M-1)=\frac{\mu X(X+1)}{KR^d}+\Theta^{1-sla}(M)\nonumber\\
&=X\left[[F(\delta(X))]^{M}g(\delta(X))+\frac{\mu (X+1)}{KR^d}\right].
\end{align}
Since $g(\delta(X))<0$ and $0<F(\delta(X))<1$, $[F(\delta(X))]^{M}g(\delta(X))+\frac{\mu (X+1)}{KR^d}$ increases with $M$. Accordingly, if $\Delta^{1-sla}(N)<0$, i.e., $[F(\delta(X))]^{N}g(\delta(X))+\frac{\mu (X+1)}{KR^d}<0$, $Z(M)$ decreases with $M$ and we have $M^*=N$. If $\Delta^{1-sla}(1)>0$, i.e., $[F(\delta(X))]g(\delta(X))+\frac{\mu (X+1)}{KR^d}>0$, $Z(M)$ increases with $M$, and thus $M^*=0$.

Otherwise, the root of $\Delta^{1-sla}(M)=0$ can be calculated as $M^*=\left\langle\log_{F(\delta(X))}\left(\frac{\mu (X+1)}{-KR^dg(\delta(X))}\right)\right\rangle$. We have $\Delta^{1-sla}(M)<0$ when $M\leq M^*$ and $\Delta^{1-sla}(M)>0$ when $M>M^*$. That is, $Z(M)$ decreases with $M$ when $M\leq M^*$ and $Z(M)$ increases with $M$ when $M>M^*$. In this case, the optimal number of downloaded layers is $\left\langle\log_{F(\delta(X))}\left(\frac{\mu (X+1)}{-KR^dg(\delta(X))}\right)\right\rangle$.
%Then,
%%
%\begin{align}\label{second_order}
%\Gamma^{1-sla}(\tilde{N})&=\Delta^{1-sla}(\tilde{N})-\Delta^{1-sla}(\tilde{N}-1)=\Lambda(\tilde{N})+\frac{\mu(\nu-1) I_{\tilde{N}-1}}{K}\nonumber\\
%&=I_{\tilde{N}-1}\left[[F(\delta(\nu))]^{\tilde{N}-1}g(\delta(\nu))\left[\nu F(\delta(\nu)) -1\right]+\frac{\mu(\nu-1)}{K}\right], \tilde{N}>2.
%\end{align}
%%
%Since $1\leq\nu<\frac{1}{F(\delta(\nu))}$ and $\Lambda(\tilde{N})>0$ according to Lemma 2, we have $\Gamma^{1-sla}(\tilde{N})>0$ and $\Delta^{1-sla}(\tilde{N})$ increases with $\tilde{N}$. If $\Delta^{1-sla}(N)<0$, $Z(\tilde{N})$ decreases with $\tilde{N}$ and we have $\tilde{N}^*=N$. If $\Delta^{1-sla}(2)>0$, $Z(\tilde{N})$ increases with $\tilde{N}$ and we have $\tilde{N}^*=1$. Otherwise, let $\Delta^{1-sla}(\tilde{N})=0$ and we have $\tilde{N}^*=\left\langle\log_{F(\delta(\nu))}\left(\frac{\mu}{-Kg(\delta(\nu))}\right)\right\rangle$.
\end{proof}

According to Proposition 4.3, we can optimize the number of downloaded layers $M$ using the closed-form expression \eqref{opt_layer} directly. %In this case, the overall computational complexity for solving Problem (P1) reduces to $O(M^*+1)$, where $M^*\in\{0,1,...,N\}$.

\subsection{Hybrid Algorithm}

Notice that it takes a linear computational complexity to solve Problem (P1) when the model splitting decision is based on the suboptimal 1-sla stopping rule. On the other hand, according to Section III C, the backward-induction based optimal model placement and splitting algorithm takes a polynomial time complexity. As an alternative to balance between solution optimality and computational complexity, we propose in this subsection a hybrid algorithm to solve the Problem (P1). In particular, we find the approximated optimal model placement decision $M^{hybrid}$, assuming the 1-sla stopping rule. That is
\begin{align}
M^{hybrid}=\arg\min_{M}Z(M,S_1(M,\bm{\gamma})).
\end{align}
Then, we fix $M^{hybrid}$ and find the optimal model splitting point $S^*(M^{hybrid},\bm{\gamma})$ using the backward induction algorithm in Section III A. Based on the complexity discussions in Section III C and Section IV B, the overall computational complexity of the hybrid algorithm is  $O(N)$. The proposed hybrid algorithm also has a linear computational complexity, while achieving a better performance compared with the 1-sla stopping rule based algorithm in Section IV B.

%The proposed hybrid algorithm has a much lower computational complexity than the backward induction based searching method in Section III C, while achieving a better performance than the 1-sla stopping rule based algorithm proposed in Section IV B.

%{\color{blue}{
%\emph{\textbf{Remark 4.1:}} For comparison, we summarize the computational complexities of different algorithms for solving the joint optimization of model placement and online model splitting decisions in Table I.
%
%\begin{table}[htb]
%\centering
%{\color{blue}{
%\begin{tabular}{cccc}
%\toprule
%Backward-induction based algorithm & $O(N(N+1)/2)$ \\
%\midrule
%1-sla based algorithm& $O(2N+1)$ \\
%\midrule
%1-sla based algorithm for the MLP with equal neurons& $O(1)-O(N+1)$ \\
%\midrule
%Hybrid algorithm& $O(2N+1)-O(3N+1)$ \\
%\bottomrule
%\end{tabular}
%\caption{Computational complexities of different algorithms for solving Problem (P1).}
%}}
%\end{table}
%}}

\section{Simulation Results}

In this section, we evaluate the proposed algorithms with different DNN architectures. Specifically, we consider the classical autoencoder and AlexNet. As shown in Fig. 5, the considered autoencoder consists of one input layer, one output layer, and seven hidden layers, where the number of neurons in the hidden layers is $\{X_i\}=\{128,~64,~32,~10,~32,~64,~128\}$. We set the number of neurons in the output layer as 784, which is the same as that in the input layer. Likewise, as shown in Fig. 6, the considered AlexNet contains eight layers: the first five are convolutional and the remaining three are fully-connected. The configuration of the AlexNet follows \cite{alexnet}, where there are 1000 class lables and the input image size is $227*227*3$.  We suppose that $\lambda=\mu=8$ Bytes, where $\lambda$ and $\mu$ are defined in \eqref{def_lamb} and \eqref{def_mu}, respectively. In addition, one floating-point multiply-add calculation requires $\alpha=100$ CPU cycles.

\begin{figure}[htb]
\begin{centering}
\includegraphics[scale=0.6]{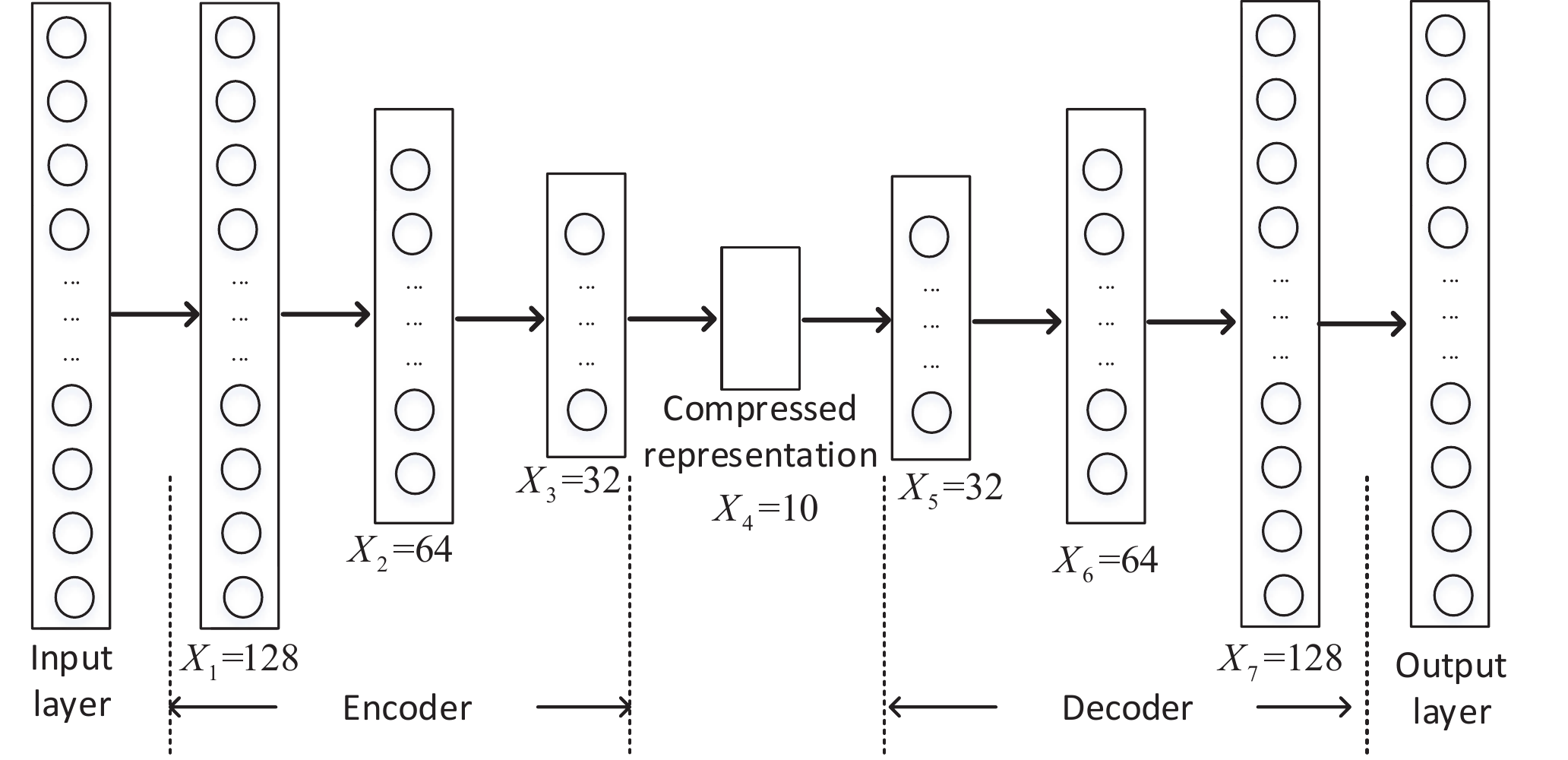}
\vspace{-0.1cm}
 \caption{The considered autoencoder structure.}
\end{centering}
\vspace{-0.1cm}
\end{figure}

\begin{figure}[htb]
\begin{centering}
\includegraphics[scale=0.65]{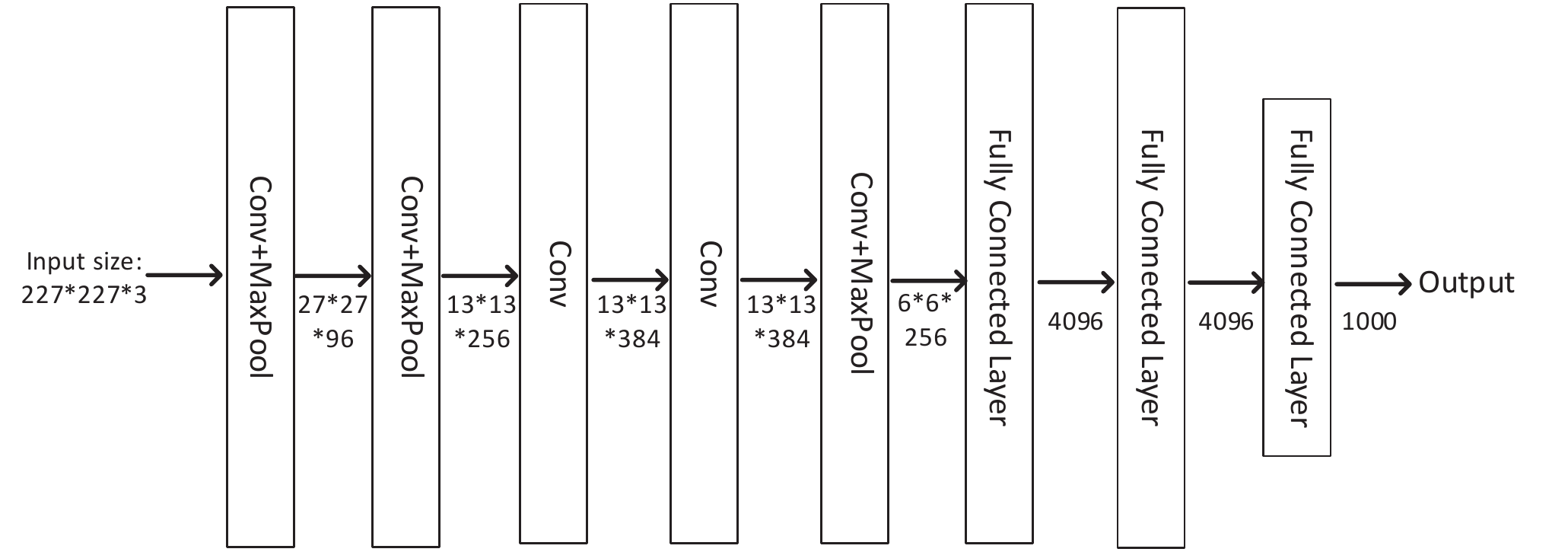}
\vspace{-0.1cm}
 \caption{The considered AlexNet structure.}
\end{centering}
\vspace{-0.1cm}
\end{figure}

The wireless channels are modeled as the Rayleigh block fading with large-scale path loss. In this case, the instantaneous SNR $\gamma_n$ is exponentially distributed with PDF
\begin{align}
f_n(\gamma_n)=\frac{1}{\bar{\gamma}_n}e^{-\frac{\gamma_n}{\bar{\gamma}_n}},
\end{align}
where $\bar{\gamma}_n=\frac{P}{\sigma^2}A_{d}(\frac{3\cdot 10^8}{4\pi f^cd})^{PL}$ is the average SNR at model splitting point $n$.
%We assume that the average channel gain $\bar{h}_{n}$ follows the free-space path loss model
%%
%\begin{align}
%\bar{h}_{n}=A_{d}(\frac{3\cdot 10^8}{4\pi f_cd})^{PL}, \forall n,
%\end{align}
%%
$A_d=4.11$ denotes the antenna gain, $f^c=915$ MHz denotes the carrier frequency, $d$ in meters denotes the distance between the WD and the BS, and $PL=3$ denotes the pass loss exponent. The transmit power of the WD and the BS is 100 mW and 1 W, respectively. The noise power $\sigma^{2}=10^{-10}$ W.  We set the computing efficiency parameter $\kappa=10^{-26}$, and the bandwidth $W=2$ MHz. The weights of energy consumption and time of the WD are set as $\beta_{t}=\beta_{e}=0.5$. The CPU frequencies at the WD and the edge server are $10^8$ and $10^{10}$ cycles/second, respectively.

\subsection{Optimality Analysis for One-Stage Look-Ahead Stopping Rule}

\begin{figure}[htb]
\begin{centering}
\includegraphics[scale=0.5]{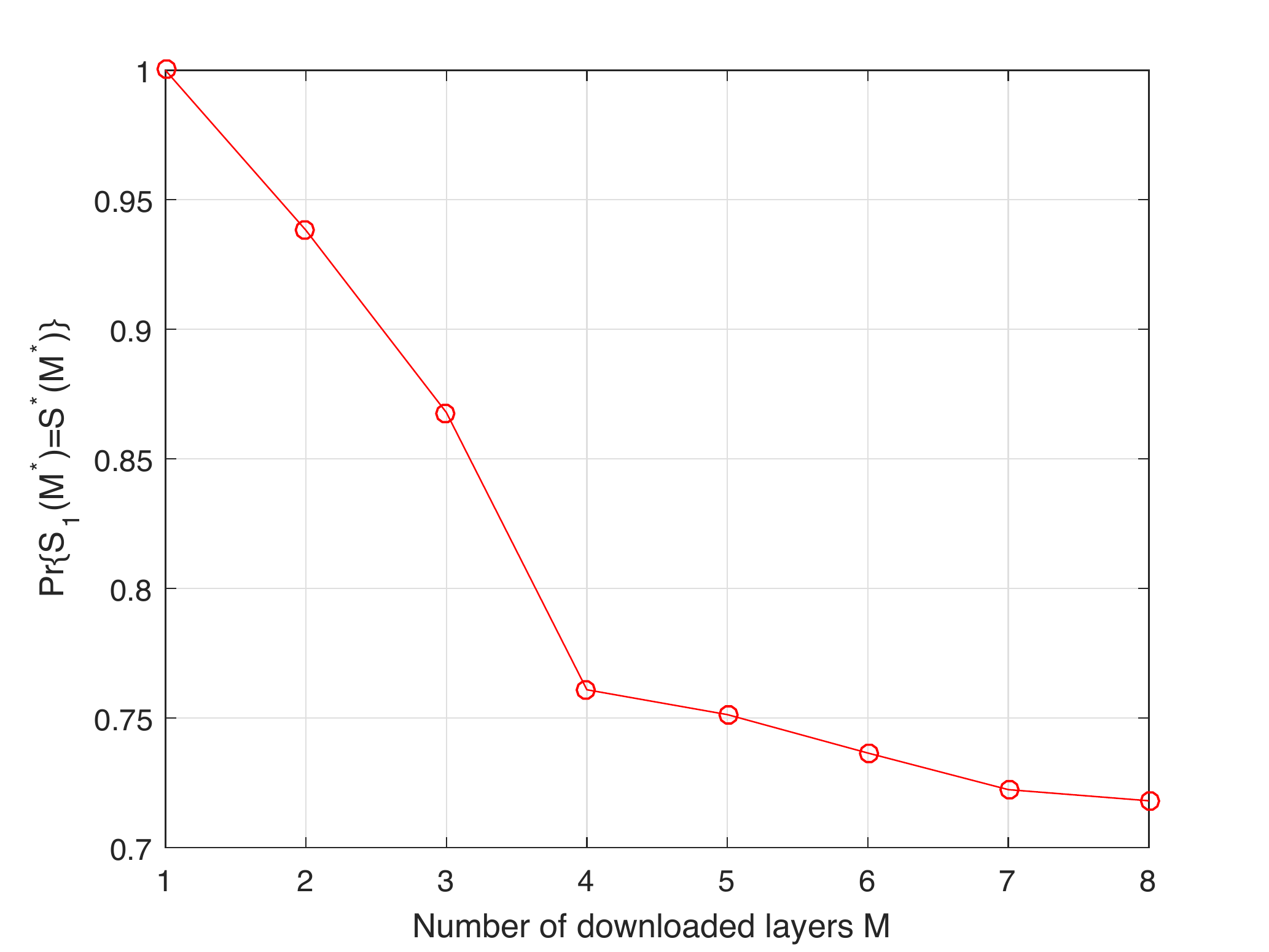}
\vspace{-0.1cm}
 \caption{Optimality probability for one-stage look-ahead rule versus the number of downloaded layers $M$ in the considered autoencoder.}
\end{centering}
\vspace{-0.1cm}
\end{figure}

\begin{figure}[htb]
\begin{centering}
\includegraphics[scale=0.5]{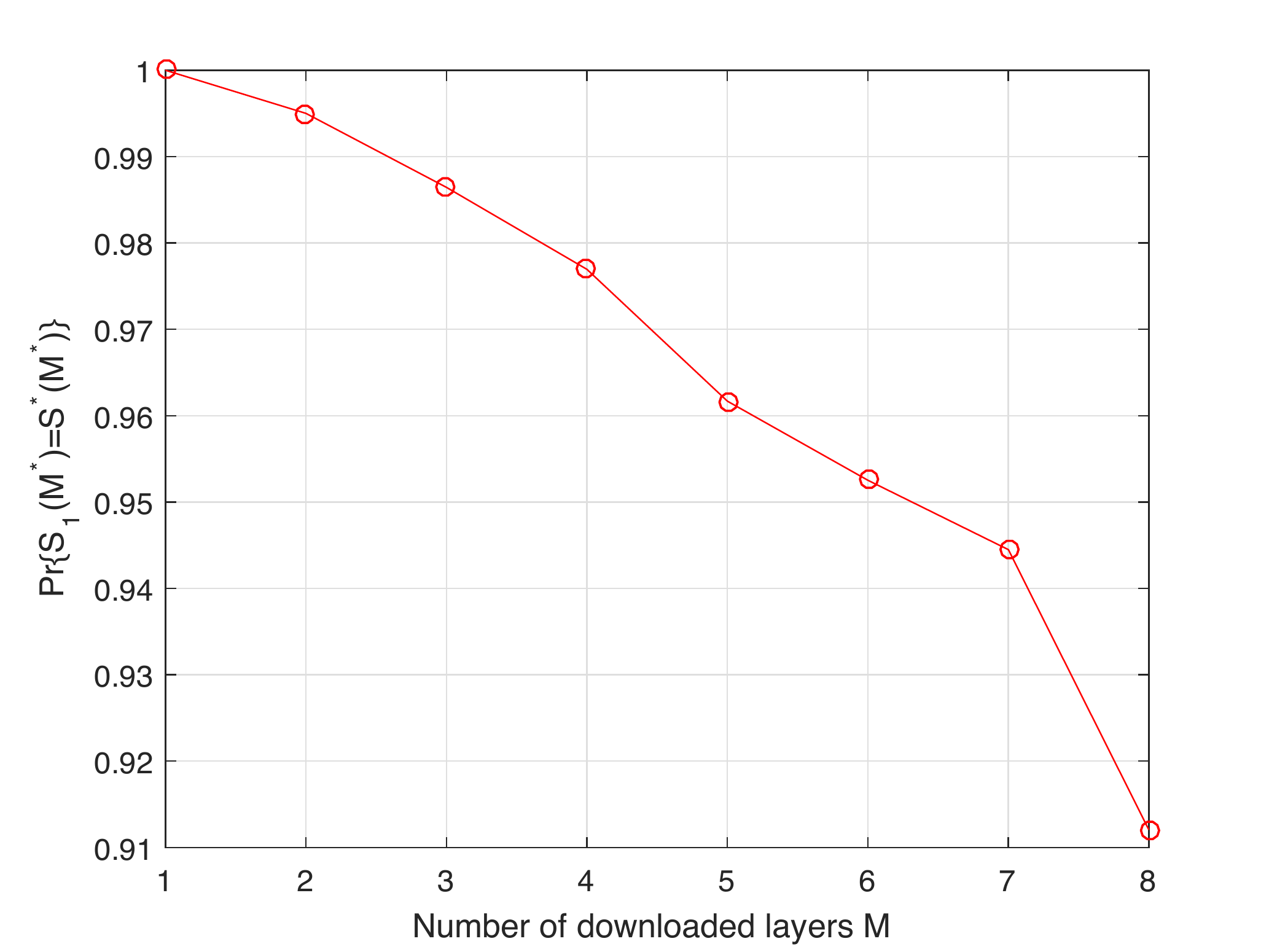}
\vspace{-0.1cm}
 \caption{Optimality probability for one-stage look-ahead rule versus the number of downloaded layers $M$ in the considered AlexNet.}
\end{centering}
\vspace{-0.1cm}
\end{figure}

In Fig. 7 and Fig. 8, we plot the optimality probability of the 1-sla stopping rule $Pr\{S_1(M,\bm{\gamma})=S^*(M,\bm{\gamma})\}$ as a function of $M$ when $d=50$ meters with the autoencoder and AlexNet, respectively. We observe that the optimality probability decreases with $M$ for both the autoencoder and AlexNet. It is because the 1-sla stopping rule becomes more myopic when the optimal stopping problem has a larger horizon (i.e., $M$ is large). It can be seen that when $M=1$, $Pr\{S_1(M,\bm{\gamma})=S^*(M,\bm{\gamma})\}=1$, which means that the 1-sla stopping rule $S_1(1,\bm{\gamma})$ is equal to the optimal $S^*(1,\bm{\gamma})$ in this case. Besides, we observe that the optimality probability is larger than 0.7 and 0.91 for any $M$ with the autoencoder and AlexNet, respectively. This confirms the effectiveness of the 1-sla stopping rule.  One interesting observation is that the optimality probability of the 1-sla stopping rule for the AlexNet is larger than that for the autoencoder. It is due to the fact that compared with the autoencoder, higher computation workloads of the early convolutional layers in the AlexNet dominate the decision making, which leads to an early stopping with higher probability for the backward-induction based optimal stopping rule. In this case, the myopic 1-sla stopping rule is close to the optimum by accounting for the expected ETC of continuing for just one stage.

\subsection{Performance Evaluation When $K=\infty$}

In this subsection, we consider the scenario where $K=\infty$, i.e., the cost of model placement is negligible. In this case, the overall average cost $Z$ is equal to the average inference cost $E_{\bm{\gamma}}[\eta_{S(M,\bm{\gamma})}(\gamma_{S(M,\bm{\gamma})})]$.

\begin{figure}[htb]
\begin{centering}
\includegraphics[scale=0.5]{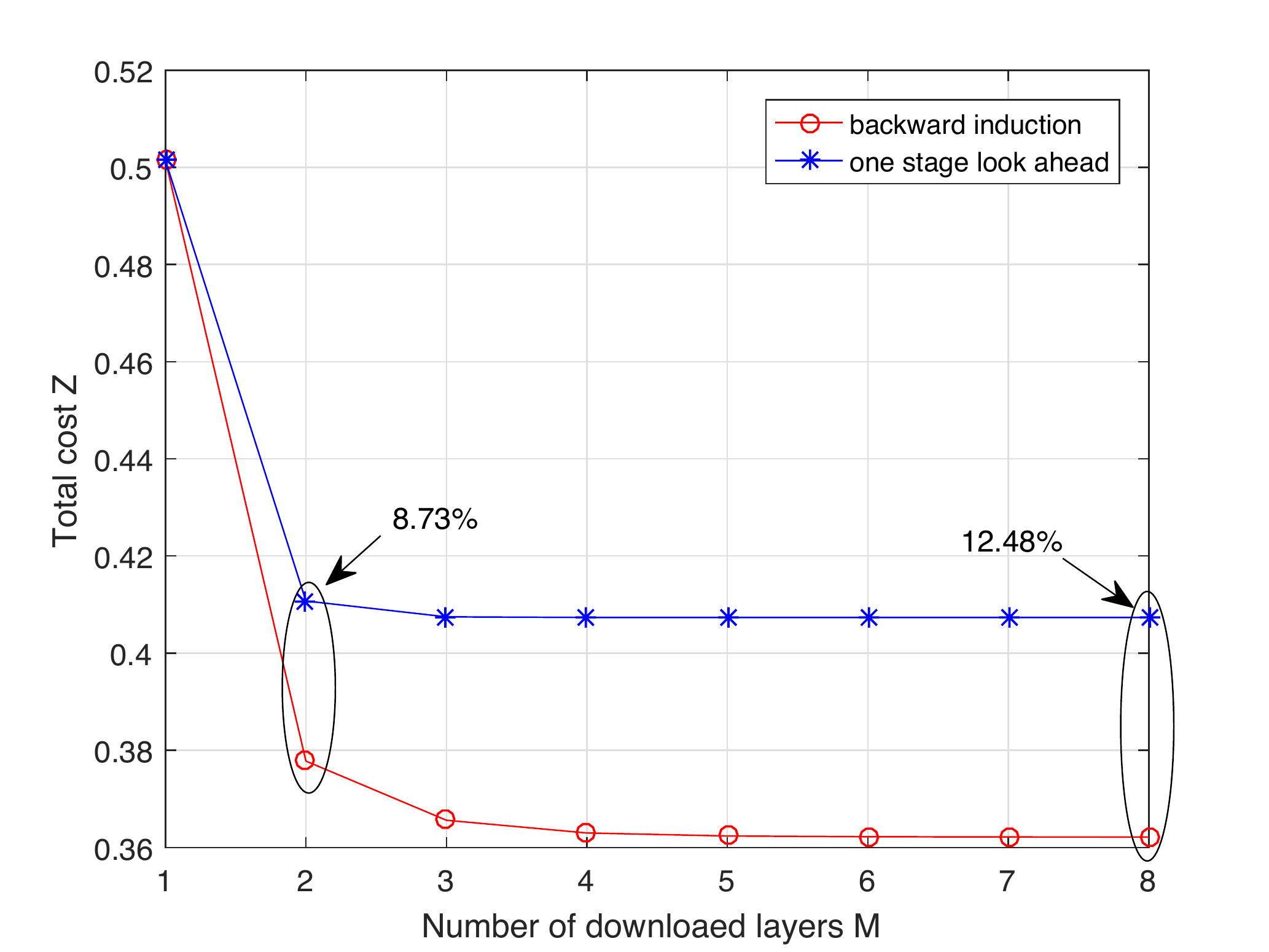}
\vspace{-0.1cm}
 \caption{Performance gap between backward induction and one-stage look-ahead rule when $K=\infty$ in the considered autoencoder.}
\end{centering}
\vspace{-0.1cm}
\end{figure}

\begin{figure}[htb]
\begin{centering}
\includegraphics[scale=0.5]{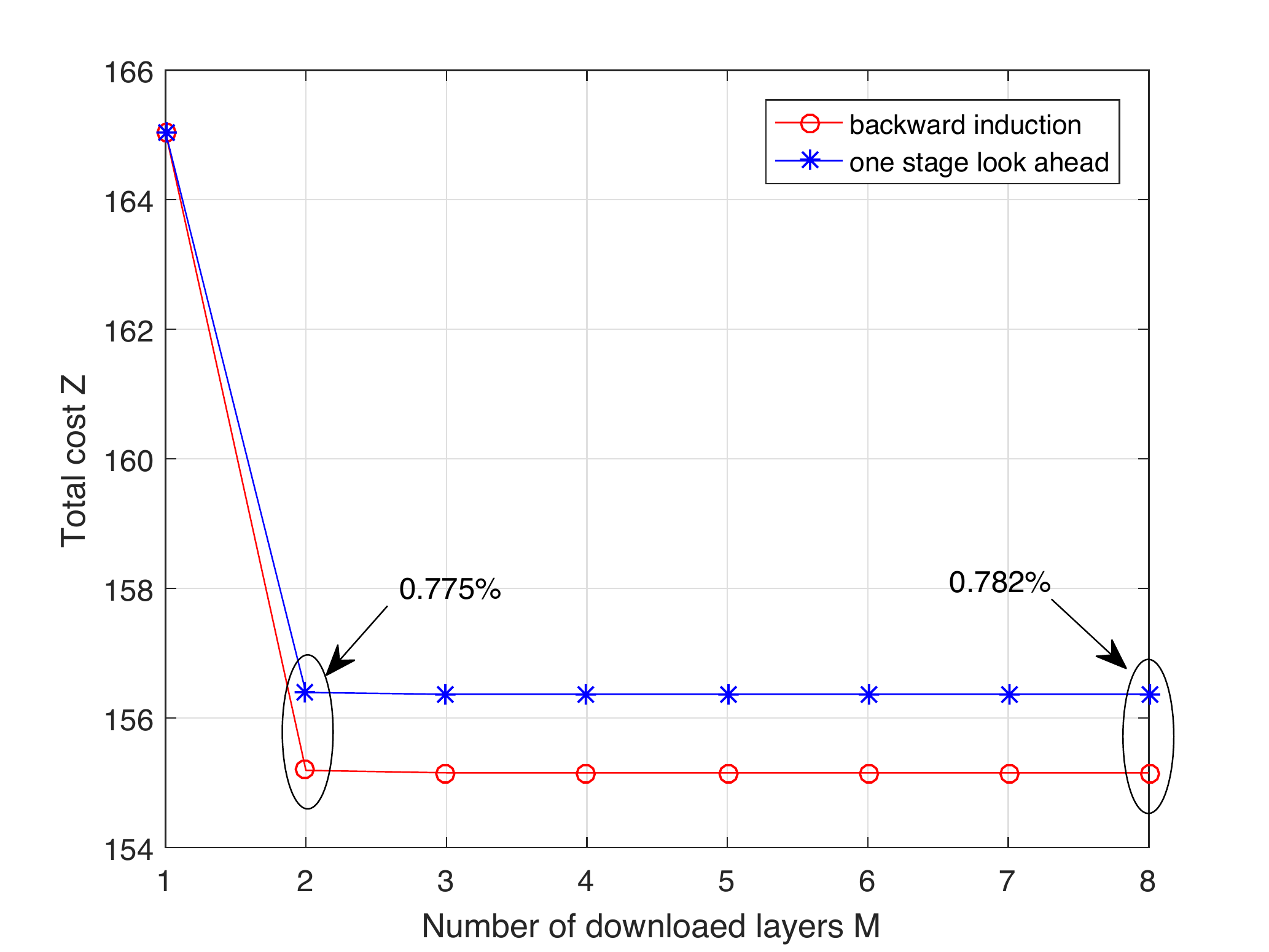}
\vspace{-0.1cm}
 \caption{Performance gap between backward induction and one-stage look-ahead rule when $K=\infty$ in the considered AlexNet.}
\end{centering}
\vspace{-0.1cm}
\end{figure}

In Fig. 9 and Fig. 10, we illustrate the total average cost $Z$ versus the number of downloaded layers $M$ when $K=\infty$ and $d=50$ meters. We observe that for both the autoencoder and AlexNet, $Z$ decreases with the increase of $M$ for both optimal (through backward induction) and 1-sla stopping rules. It is because when more layers are downloaded from the edge server, the WD has more choices of the model splitting points, which improves the average inference performance. In particular, when $M=1$, the 1-sla stopping rule has the same performance as the optimal stopping rule. In addition, the gap of $Z$ between the optimal and 1-sla stopping rules increases when $M$ grows. For example, $Z$ grows from $8.73\%$ to $12.48\%$ when $M$ increases from 2 to 8 in Fig. 9.

Compared to the autoencoder, the performance gap between the optimal and 1-sla stopping rules in the AlexNet is smaller, e.g., $0.782\%$ when $M=8$. It is because as illustrated in Fig. 7 and Fig. 8, the optimality probability of the 1-sla stopping rule for the AlexNet is always larger than that for the autoencoder. Moreover, Fig. 10 shows that the inference cost in the AlexNet remains stable as $M$ varies from 2 to 8. It is due to the fact that the heavy computation workloads of the early convolutional layers in the AlexNet dominate the decision making for the model splitting point selection. That is, even though more layers are downloaded, the WD still prefers to stop at an early stage.

%\begin{figure}[htb]
%\begin{centering}
%\includegraphics[scale=0.7]{model_example.eps}
%\vspace{-0.1cm}
% \caption{The considered topological call graph in simulation, where $\{L_i\}=[70.8 ~95.3~ ~86.4 ~18.6 ~158.6]$ (Mcycles).}
%\end{centering}
%\vspace{-0.1cm}
%\end{figure}
%
%\begin{figure}[htb]
%\begin{centering}
%\includegraphics[scale=0.7]{gap.eps}
%\vspace{-0.1cm}
% \caption{Performance gap between backward induction and one-stage look-ahead rule.}
%\end{centering}
%\vspace{-0.1cm}
%\end{figure}
%
%\begin{figure}[htb]
%\begin{centering}
%\includegraphics[scale=0.7]{prob_bi_vs_onesla.eps}
%\vspace{-0.1cm}
% \caption{Probability of stopping at each stage for backward induction and one-stage look-ahead rule using the example in Fig. 4.}
%\end{centering}
%\vspace{-0.1cm}
%\end{figure}
%
%\begin{figure}[htb]
%\begin{centering}
%\includegraphics[scale=0.7]{prob_bi_diff.eps}
%\vspace{-0.1cm}
% \caption{Probability of stopping at each stage for backward induction under different number of downloaded layers $M$ using the example in Fig. 4.}
%\end{centering}
%\vspace{-0.1cm}
%\end{figure}
%
%\begin{figure}[htb]
%\begin{centering}
%\includegraphics[scale=0.7]{distance_Z.eps}
%\vspace{-0.1cm}
% \caption{Total cost versus the distance between the WD and edge server when $K=10$ using the example in Fig. 4.}
%\end{centering}
%\vspace{-0.1cm}
%\end{figure}
%
%\begin{figure}[htb]
%\begin{centering}
%\includegraphics[scale=0.7]{distance_M.eps}
%\vspace{-0.1cm}
% \caption{The number of downloaded layers $M$ versus the distance between the WD and edge server using the example in Fig. 4.}
%\end{centering}
%\vspace{-0.1cm}
%\end{figure}

\subsection{Performance Evaluation When $K<\infty$}

We now consider the general scenario where the DNN parameters need to be updated from time to time, i.e., $K<\infty$.

\begin{figure}[htb]
\begin{centering}
\includegraphics[scale=0.5]{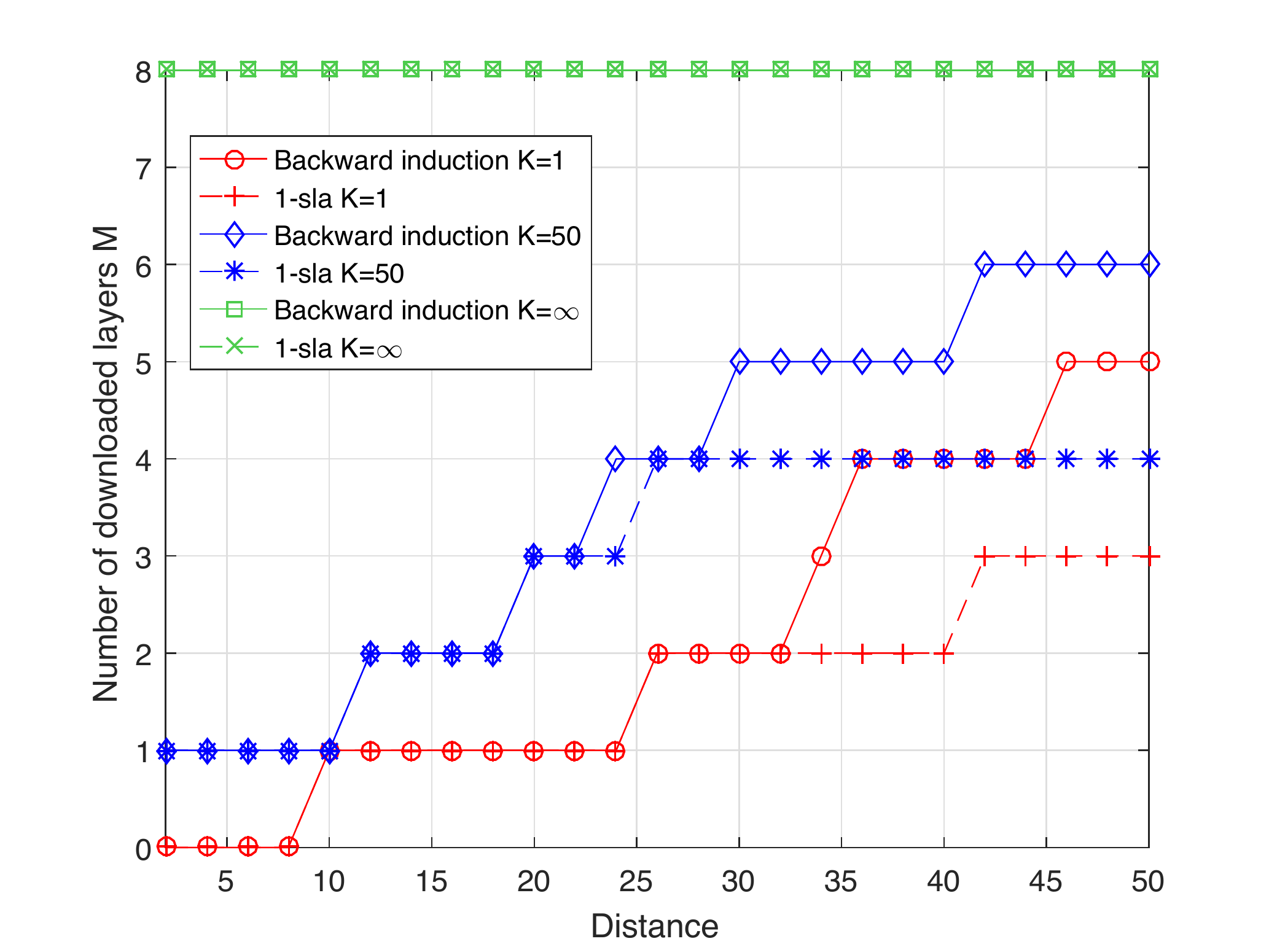}
\vspace{-0.1cm}
 \caption{The number of downloaded layers $M$ versus the distance between the WD and edge server in the considered autoencoder.}
\end{centering}
\vspace{-0.1cm}
\end{figure}

\begin{figure}[htb]
\begin{centering}
\includegraphics[scale=0.5]{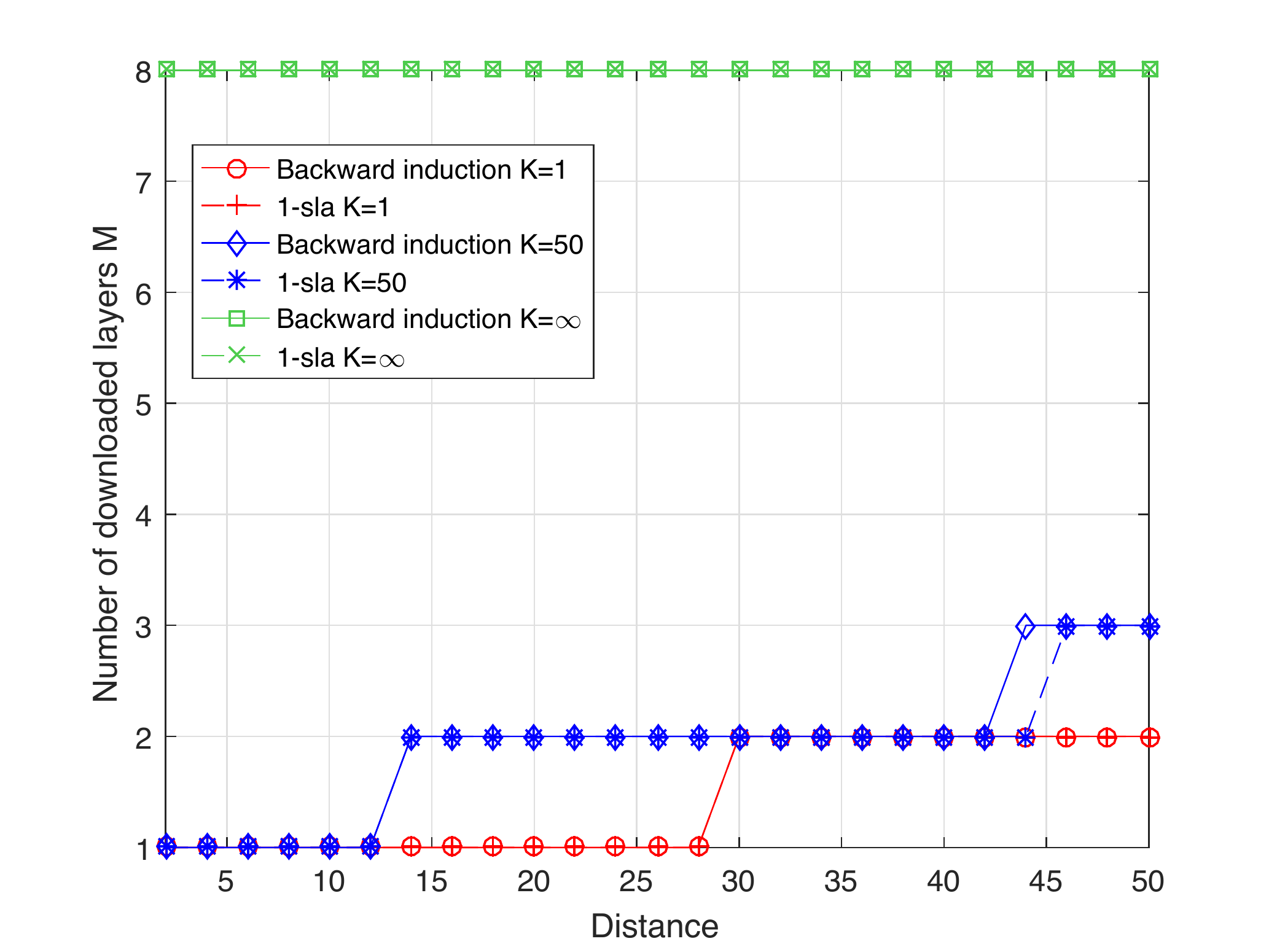}
\vspace{-0.1cm}
 \caption{The number of downloaded layers $M$ versus the distance between the WD and edge server in the considered AlexNet.}
\end{centering}
\vspace{-0.1cm}
\end{figure}

In Fig. 11, we plot the optimal $M^*$ as a function of the distance $d$ under different model update frequencies $K$ with the autoencoder. It is observed that the optimal $M^*$ increases with the distance $d$. This is because a larger distance leads to a higher uplink offloading cost for the intermediate feature. In this case, the average inference cost can be improved by downloading more layers to the WD. Besides, we observe that the optimal $M^*$ increases with the model update frequency parameter $K$, which means that the BS tends to download more layers to the WD when each downloaded model can be used for more inference requests. In particular, when $K=\infty$, all the layers of the DNN are downloaded, i.e., $M^*=N$. Moreover, given $K$, the optimal $M^*$ under 1-sla stopping rule is not larger than that under the optimal backward induction rule due to the myopic property of the 1-sla stopping rule.

In Fig. 12, we demonstrate the impact of distance $d$ on the optimal $M^*$ in the AlexNet. Comparing with Fig. 11, we can draw two conclusions from Fig. 12. First, the BS tends to download smaller number of layers to the WD for the AlexNet. This is due to the larger computation workloads and higher model parameter sizes of the convolutional layers in the AlexNet compared with the autoencoder. This leads to higher costs for the local computing and model parameter downloading. Second, the optimal $M^*$ computed based on the 1-sla rule is close to that computed based on the backward induction rule. It is because according to Fig. 7 and 8, the 1-sla stopping rule has higher optimality probability in the AlexNet compared with the autoencoder.
%the WD calls for an early stopping with higher probability due to the large computation workloads of the convolutional layers in the AlexNet.

\begin{figure}[htb]
\begin{centering}
\includegraphics[scale=0.5]{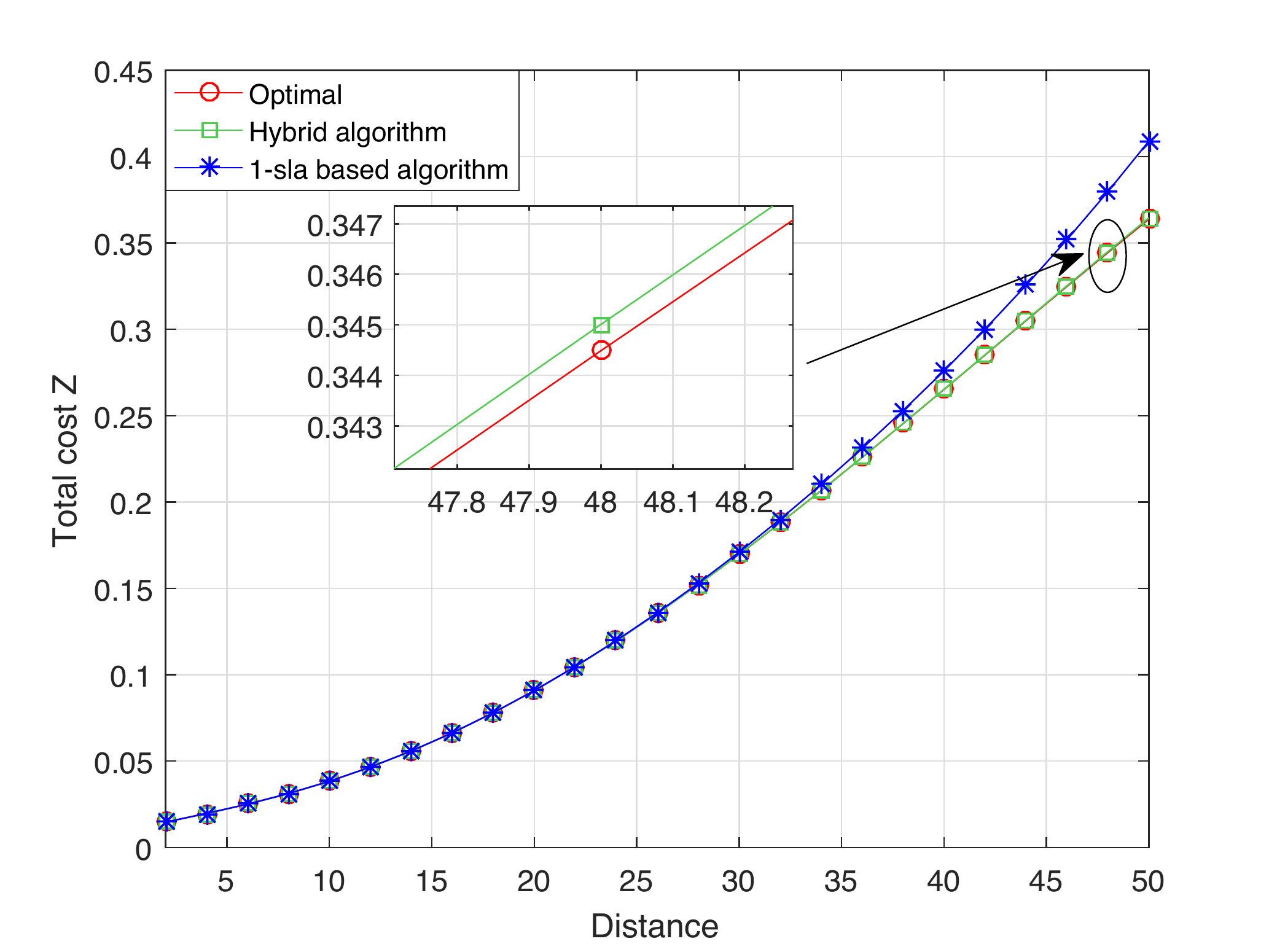}
\vspace{-0.1cm}
 \caption{Total cost versus the distance between the WD and edge server when $K=50$ in the considered autoencoder.}
\end{centering}
\vspace{-0.1cm}
\end{figure}

\begin{figure}[htb]
\begin{centering}
\includegraphics[scale=0.5]{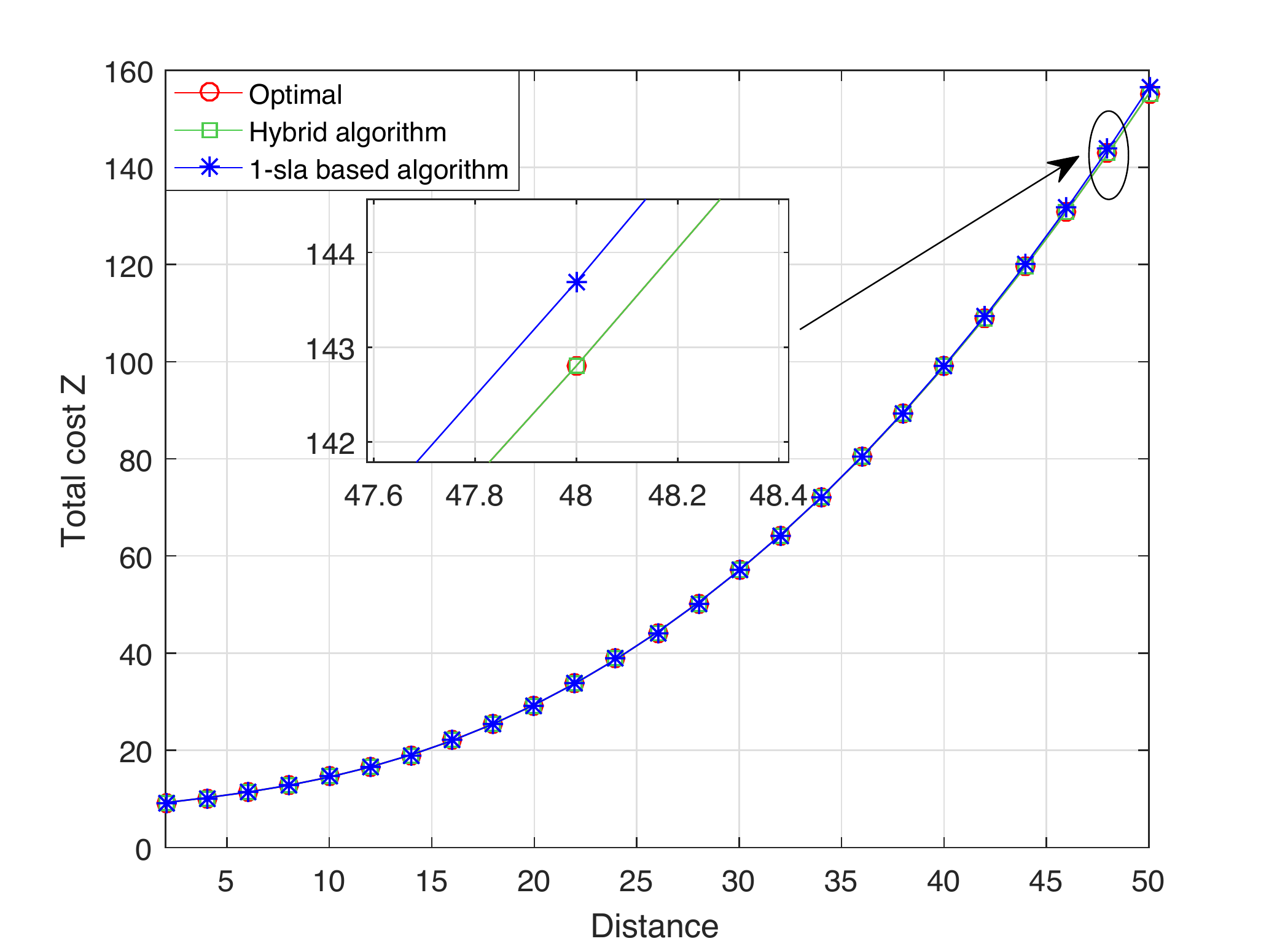}
\vspace{-0.1cm}
 \caption{Total cost versus the distance between the WD and edge server when $K=50$ in the considered AlexNet.}
\end{centering}
\vspace{-0.1cm}
\end{figure}

In Fig. 13 and Fig. 14, we plot the total average cost $Z$ when the distance $d$ varies and $K=50$. Fig. 13 shows that the hybrid algorithm outperforms the 1-sla stopping rule based scheme and performs very closely to the optimal solution.  Jointly considering the close-to-optimal performance and the complexity of the hybrid algorithm, it is a preferred choice for practical implementation.
%, e.g., $0.2\%$ and $12.4\%$ higher total average costs than the optimum when $d=50$ meters for the hybrid and 1-sla stopping rule based algorithms in the autoencoder, respectively.
One interesting observation is that when the distance is small (e.g., below 30 meters for the autoencoder), the 1-sla rule based algorithm is also close to the optimum. This is because a smaller distance leads to a smaller optimal $M^*$ as illustrated in Fig. 11 and 12. Under a smaller $M^*$, the 1-sla stopping rule has a higher optimality probability. Likewise, for the AlexNet in Fig. 14, the 1-sla stopping rule based scheme and the hybrid algorithm achieve close-to-optimal performances even when $d$ is large.

%\subsection{Complexity of the Proposed Algorithms}
%
%\begin{table*}[htb]
%\centering
%\begin{tabular}{cccc}
%\toprule
%  &Optimal (Backward Induction Based) Algorithm & 1-sla Based Algorithm & Hybrid Algorithm \\
%\midrule
% Autoencoder&0.1573 s & 0.1291 s & 0.1440 s  \\
%\midrule
%AlexNet&0.1667 s& 0.1274 s & 0.1432 s  \\
%
%
%\bottomrule
%\end{tabular}
%\caption{Comparisons of computation time under different algorithms.}
%\end{table*}
%
%At last, we compare the computational complexity among the proposed algorithms, where the distance $d=50$ meters and the model update frequency $K=50$. As shown in Table I, the 1-sla based and hybrid algorithms require shorter runtime than the backward induction based algorithm. In particular, for the AlexNet, around $23.58\%$ and $14.1\%$ lower average runtime is achieved by the 1-sla based and hybrid algorithms, respectively. Therefore, the hybrid algorithm can achieve the near-optimal performance in compensation for a slight increase of the computational complexity compared with the 1-sla based algorithm.

\section{Conclusions}

This paper has studied the joint optimization of model placement and online model splitting strategy to minimize the expected energy and time cost of device-edge co-inference. Given the model placement decision, we have formulated the problem of online model splitting as an optimal stopping problem with finite horizon. Then, we have proposed an online algorithm based on backward induction to find the optimal stopping rule (i.e., the optimal model splitting strategy). To simplify the analysis, we have proposed a 1-sla stopping rule for model splitting, based on which an efficient algorithm is proposed to optimize the model placement decision. Under a specific DNN structure, we have derived the closed-form expressions for the 1-sla stopping rule based model placement. We have further proposed a hybrid algorithm to balance between the solution optimality and the computational complexity. Simulation results have validated the benefits of jointly considering the model placement and splitting under various DNN structures.

\begin{footnotesize}
\bibliographystyle{IEEEtran}

\begin{thebibliography}{10}
\providecommand{\url}[1]{#1}
\csname url@samestyle\endcsname
\providecommand{\newblock}{\relax}
\providecommand{\bibinfo}[2]{#2}
\providecommand{\BIBentrySTDinterwordspacing}{\spaceskip=0pt\relax}
\providecommand{\BIBentryALTinterwordstretchfactor}{4}
\providecommand{\BIBentryALTinterwordspacing}{\spaceskip=\fontdimen2\font plus
\BIBentryALTinterwordstretchfactor\fontdimen3\font minus
  \fontdimen4\font\relax}
\providecommand{\BIBforeignlanguage}[2]{{%
\expandafter\ifx\csname l@#1\endcsname\relax
\typeout{** WARNING: IEEEtran.bst: No hyphenation pattern has been}%
\typeout{** loaded for the language `#1'. Using the pattern for}%
\typeout{** the default language instead.}%
\else
\language=\csname l@#1\endcsname
\fi
#2}}
\providecommand{\BIBdecl}{\relax}
\BIBdecl

\bibitem{eisurvey}
Z.~{Zhou}, X.~{Chen}, E.~{Li}, L.~{Zeng}, K.~{Luo}, and J.~{Zhang}, ``Edge
  intelligence: Paving the last mile of artificial intelligence with edge
  computing,'' \emph{Proc. {IEEE}}, vol. 107, no.~8, pp. 1738--1762, 2019.

\bibitem{ai3}
M.~{Chen}, U.~{Challita}, W.~{Saad}, C.~{Yin}, and M.~{Debbah}, ``Artificial
  neural networks-based machine learning for wireless networks: A tutorial,''
  \emph{IEEE Communications Surveys Tutorials}, vol.~21, no.~4, pp. 3039--3071,
  2019.

\bibitem{ai1}
R.~Collobert and J.~Weston, ``A unified architecture for natural language
  processing: Deep neural networks with multitask learning,'' in
  \emph{International Conference on Machine Learning}, 2008, pp. 160--167.

\bibitem{ai2}
R.~Szeliski, \emph{Computer vision: algorithms and applications}.\hskip 1em
  plus 0.5em minus 0.4em\relax Springer Science \& Business Media, 2010.

\bibitem{ondevice1}
N.~D. {Lane}, S.~{Bhattacharya}, P.~{Georgiev}, C.~{Forlivesi}, L.~{Jiao},
  L.~{Qendro}, and F.~{Kawsar}, ``Deepx: A software accelerator for low-power
  deep learning inference on mobile devices,'' in \emph{2016 15th ACM/IEEE
  International Conference on Information Processing in Sensor Networks
  (IPSN)}, 2016, pp. 1--12.

\bibitem{ondevice2}
S.~Liu, Y.~Lin, Z.~Zhou, K.~Nan, H.~Liu, and J.~Du, ``On-demand deep model
  compression for mobile devices: A usage-driven model selection framework,''
  in \emph{Proc. 16th Annu. Int. Conf. Mobile Syst., Appl., Services}, 2018,
  pp. 389--400.

\bibitem{Bi_service}
S.~Bi, L.~Huang, and Y.-J.~A. Zhang, ``Joint optimization of service caching
  placement and computation offloading in mobile edge computing systems,''
  \emph{IEEE Transactions on Wireless Communications}, vol.~19, no.~7, pp.
  4947--4963, 2020.

\bibitem{Jia_pricing}
J.~Yan, S.~Bi, L.~Duan, and Y.-J.~A. Zhang, ``Pricing-driven service caching
  and task offloading in mobile edge computing,'' \emph{IEEE Transactions on
  Wireless Communications}, pp. 1--1, 2021.

\bibitem{Zehong}
Z.~Lin, S.~Bi, and Y.-J.~A. Zhang, ``Optimizing ai service placement and
  resource allocation in mobile edge intelligence systems,'' \emph{arXiv
  preprint arXiv:2011.05708}, 2020.

\bibitem{serverbased1}
A.~I. {Maqueda}, A.~{Loquercio}, G.~{Gallego}, N.~{Garc¨ªa}, and
  D.~{Scaramuzza}, ``Event-based vision meets deep learning on steering
  prediction for self-driving cars,'' in \emph{2018 IEEE/CVF Conference on
  Computer Vision and Pattern Recognition}, 2018, pp. 5419--5427.

\bibitem{serverbased2}
L.~Liu, H.~Li, and M.~Gruteser, ``Edge assisted real-time object detection for
  mobile augmented reality,'' in \emph{The 25th Annual International Conference
  on Mobile Computing and Networking}, 2019.

\bibitem{ei1}
H.~{Li}, C.~{Hu}, J.~{Jiang}, Z.~{Wang}, Y.~{Wen}, and W.~{Zhu}, ``Jalad: Joint
  accuracy-and latency-aware deep structure decoupling for edge-cloud
  execution,'' in \emph{2018 IEEE 24th International Conference on Parallel and
  Distributed Systems (ICPADS)}, 2018, pp. 671--678.

\bibitem{ei2}
J.~{Shao} and J.~{Zhang}, ``Bottlenet++: An end-to-end approach for feature
  compression in device-edge co-inference systems,'' in \emph{2020 IEEE
  International Conference on Communications Workshops (ICC Workshops)}, 2020,
  pp. 1--6.

\bibitem{ei3}
------, ``Communication-computation trade-off in resource-constrained edge
  inference,'' \emph{IEEE Communications Magazine}, vol.~58, no.~12, pp.
  20--26, 2020.

\bibitem{ei4}
Y.~Kang, J.~Hauswald, C.~Gao, A.~Rovinski, T.~Mudge, J.~Mars, and L.~Tang,
  ``Neurosurgeon: Collaborative intelligence between the cloud and mobile
  edge,'' in \emph{ACM SIGARCH Computer Architecture News}, vol.~45, no.~1,
  2017, pp. 615--629.

\bibitem{ei5}
J.~H. {Ko}, T.~{Na}, M.~F. {Amir}, and S.~{Mukhopadhyay}, ``Edge-host
  partitioning of deep neural networks with feature space encoding for
  resource-constrained internet-of-things platforms,'' in \emph{2018 15th IEEE
  International Conference on Advanced Video and Signal Based Surveillance
  (AVSS)}, 2018, pp. 1--6.

\bibitem{ei6}
W.~Shi, Y.~Hou, S.~Zhou, Z.~Niu, Y.~Zhang, and L.~Geng, ``Improving device-edge
  cooperative inference of deep learning via 2-step pruning,'' \emph{arXiv
  preprint arXiv:1903.03472}, 2019.

\bibitem{alexnet}
A.~Krizhevsky, I.~Sutskever, and G.~E. Hinton, ``Imagenet classification with
  deep convolutional neural networks,'' in \emph{Advances in neural information
  processing systems}, 2012, pp. 1097--1105.

\bibitem{stopping}
T.~S. Ferguson, \emph{Optimal stopping and applications}.\hskip 1em plus 0.5em
  minus 0.4em\relax http://www.math.ucla.edu/ tom/Stopping/Contents.html.

\bibitem{stopping1}
Y.~J. {Zhang}, ``Multi-round contention in wireless lans with multipacket
  reception,'' \emph{{IEEE} Trans. Wireless Commun.}, vol.~9, no.~4, pp.
  1503--1513, 2010.

\bibitem{stopping2}
Q.~{Gu}, Y.~{Jian}, G.~{Wang}, R.~{Fan}, H.~{Jiang}, and Z.~{Zhong}, ``Mobile
  edge computing via wireless power transfer over multiple fading blocks: An
  optimal stopping approach,'' \emph{{IEEE} Trans. Veh. Technol.}, vol.~69,
  no.~9, pp. 10\,348--10\,361, 2020.

\bibitem{stopping3}
J.~{Jia}, Q.~{Zhang}, and X.~S. {Shen}, ``Hc-mac: A hardware-constrained
  cognitive mac for efficient spectrum management,'' \emph{IEEE Journal on
  Selected Areas in Communications}, vol.~26, no.~1, pp. 106--117, 2008.

\bibitem{mlp}
D.~W. {Ruck}, S.~K. {Rogers}, M.~{Kabrisky}, M.~E. {Oxley}, and B.~W. {Suter},
  ``The multilayer perceptron as an approximation to a bayes optimal
  discriminant function,'' \emph{{IEEE} Trans. Neural Netw.}, vol.~1, no.~4,
  pp. 296--298, 1990.

\end{thebibliography}

\end{footnotesize}

\end{document}